\documentclass[letterpaper, journal]{IEEEtran} 
\usepackage[english]{babel}
\usepackage[T1]{fontenc}
\usepackage[nolist]{acronym}
\usepackage[ruled,vlined]{algorithm2e}
\usepackage[cmex10]{amsmath}
\usepackage{amssymb}
\usepackage{bm}
\usepackage{booktabs}
\usepackage{color}
\usepackage{float}
\usepackage{graphicx,subfigure}
\usepackage{hyperref}
\usepackage{import}
\usepackage{mathtools}
\usepackage{multirow}
\usepackage{outline}
\usepackage{paralist}
\usepackage{stmaryrd}
\usepackage{systeme}
\usepackage{units}
\usepackage{url}
\usepackage{tikz}
\usepackage{empheq}
\usepackage{booktabs}
\usepackage{subfigure}
\usetikzlibrary{tikzmark}
\usepackage{eso-pic}

\newcommand{\mvec}[1]{\boldsymbol{#1}}

\newcommand{\mrb}[1]{\left( #1 \right)} %enclose in round brackets
\newcommand{\skewSym}[1]{\ensuremath{\,\boldsymbol{S}\!\mrb{#1}}}

\newcommand{\measAngVel}{\hat{\mvec{\omega}}}

\newcommand{\topnote}[1]{%
  \AddToShipoutPictureBG*{%
    \begin{tikzpicture}[remember picture,overlay]
      \node[anchor=north west,inner sep=3pt] at ([xshift=15pt,yshift=-15pt]current page.north west) {%
        \parbox[t][0pt]{0.7\textwidth}{\raggedright\color{blue}\fontsize{8}{8}\selectfont #1}};
    \end{tikzpicture}%
  }
}

\title{Design and control of a collision-resilient aerial vehicle with an icosahedron tensegrity structure}
\author{Jiaming Zha, Xiangyu Wu, Ryan Dimick, and Mark W. Mueller
\thanks{The authors are with the HiPeRLab, University of California, Berkeley, CA 94720, USA. {\tt\small \{jiaming\_zha,wuxiangyu,rdimick,{\break}mwm\}@berkeley.edu} }
}

\begin{document}

\maketitle
\begin{abstract}
We introduce collision-resilient aerial vehicles with icosahedron tensegrity structures, capable of surviving high-speed impacts and resuming operations post-collision. We present a model-based design approach, which guides the selection of the tensegrity components by predicting structural stresses through a dynamics simulation. 
Furthermore, we develop an autonomous re-orientation controller that facilitates post-collision flight resumption. The controller enables the vehicles to rotate from an arbitrary orientation on the ground for takeoff. With collision resilience and re-orientation ability, the tensegrity aerial vehicles can operate in cluttered environments without complex collision-avoidance strategies.
These capabilities are validated by a test of an experimental vehicle operating autonomously in a previously-unknown forest environment.
\end{abstract}

\topnote{This work has been submitted to the IEEE/ASME Transactions on Mechatronics for possible publication. Copyright may be transferred without notice, after which this version may no longer be accessible.}

\section{Introduction}
\label{sec:introduction}
Autonomous aerial vehicles, being weight-sensitive, are often fragile. Damage to their propellers or electronics can result in the loss of their ability to fly. Protective measures for these vehicles typically fall into two categories:  detecting and avoiding collisions, and/or preventing physical damage caused by collisions. 
Methods of the first category focus on sensing surrounding spaces and finding safe paths based on the collected information. A survey summarizing recent development in the area is in \cite{yasin2020unmanned}. Methods of the second category, which this work belongs to, help aerial vehicles operate more safely in cluttered environments, where accidental collisions may occur due to imperfect sensing or control. In this paper, we present the design of collision-resilient flying robots, termed tensegrity aerial vehicles, featuring icosahedron tensegrity structures. The tensegrity aerial vehicles can withstand high-speed impacts and resume operation after collisions. With these capabilities, they can safely operate in cluttered environments without complex collision-avoidance strategies.

\subsection{Related work: collision-resilient flying robots}
Several approaches to create collision-resilient flying robots exist in literature. These include protecting the robot with external structures, constructing the vehicle with soft materials or morphing structures that can absorb substantial energy before breaking, or combining both approaches.
The first approach focuses on shielding vulnerable parts from obstacles. Examples include protective structures such as propeller guards \cite{salaan2019development}, spherical body shells \cite{briod2014collision, elios3}, cylindrical body guards \cite{jia2022quadrotor}, free-to-rotate origami shells \cite{sareh2018rotorigami} and protectors with mortise-and-tenon structures \cite{de2021resilient}. The second approach utilizes materials or parts designed to absorb significant energy before breaking. Examples include dual-stiffness frames that soften under impact to avoid damage \cite{mintchev2018bioinspired}, flexible rotor blades that can bend without breaking during collisions \cite{jang2019design}, and passively-foldable airframes \cite{shu2019quadrotor}. Some designs integrate both approaches, such as propeller-guarded vehicles with spring-loaded arms \cite{liu2021toward}, and quadcopters with passively-morphing exoskeletons \cite{de2022being}.
 
\begin{figure}[b]
    \centering
    \includegraphics[width=0.45\linewidth]{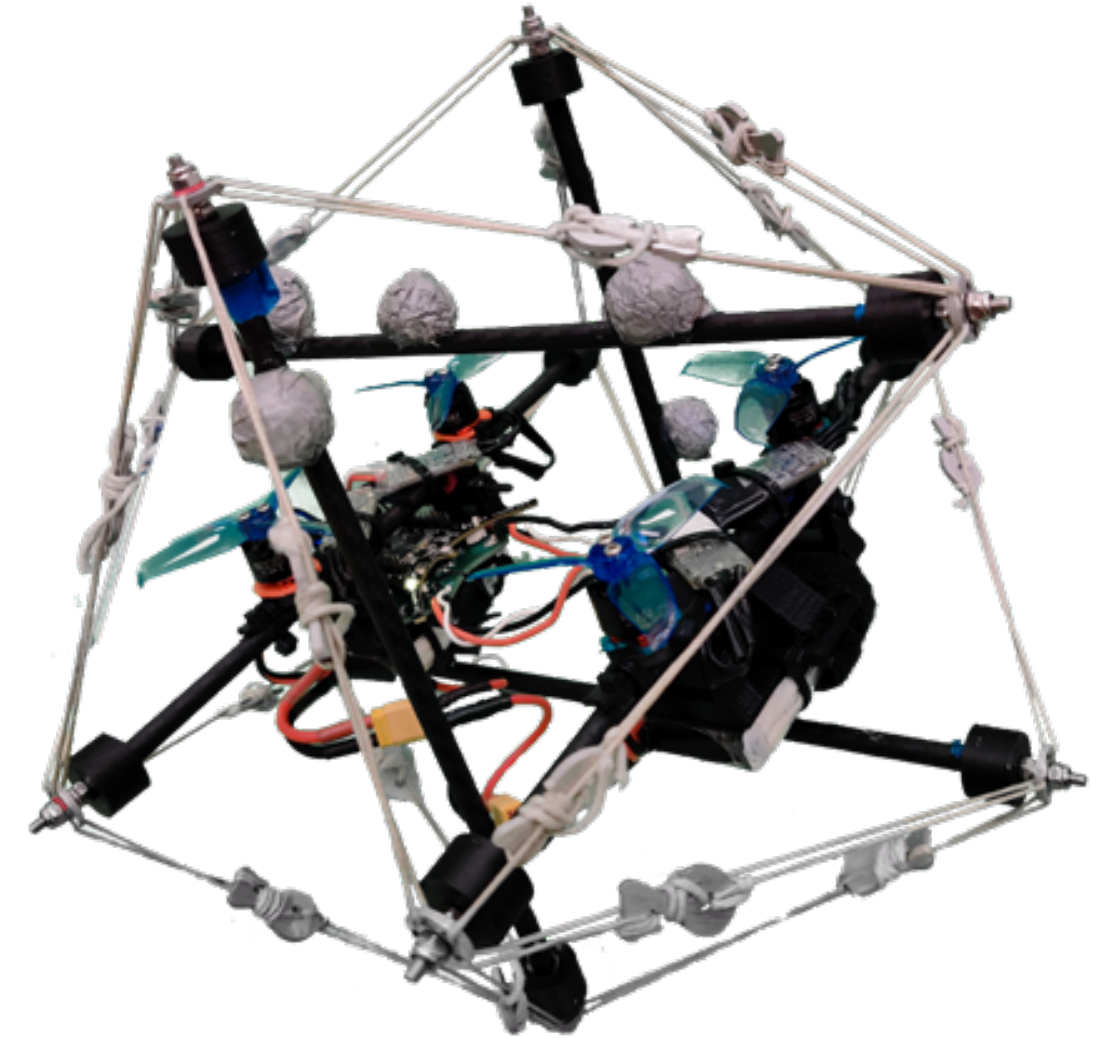}
    \caption{The icosahedron tensegrity aerial vehicle created with the proposed model-based design approach. The length of each rod in the shell is 20cm. All electronics are directly mounted on the tensegrity rods.}
    % It carries a wireless camera used for surveying the surroundings.}
    \label{VehiclePic}
\end{figure}
\subsection{Related work: tensegrity structure}
Tensegrity structures have gained popularity in recent years for their collision-resilience. Comprised of rigid bodies suspended in a tension network, tensegrity structures can distribute external loads among structural members through tension and compression, effectively avoiding large stress concentrations caused by bending. Due to their structural advantages, tensegrity structures have been proposed for applications in diverse areas such as aircraft wings \cite{9351657}, landers \cite{rimoli2016impact,zhang2021orientation, garanger2020soft}, exploratory rovers \cite{sunspiral2013tensegrity,kim2016hopping}, swarm terrestrial explorers \cite{8460667}, and general collision-resilient robotic platforms \cite{zappetti2022dual}. The benefits of tensegrities also make them suitable for aerial vehicles. An investigation comparing different tensegrity shells for aerial vehicles, supported by drop tests, is detailed in \cite{tensDrone}. It concludes that for tensegrities of the same size, those composed of lighter and stiffer materials can withstand higher drops. Our previous work \cite{zha2020collision} presented a quadcopter design with a stiff tensegrity shell. Another example is the `Tensodrone' which incorporates a soft tensegrity shell with springs, as showcased in \cite{savin2022mixed}, along with a subsequent design featuring self-morphing abilities. The soft tensegrity design helps increase collision resilience at the potential cost of larger vehicle size and vibration. The comparison between stiff and soft tensegrity shell designs is further discussed in Section \ref{sec:tensegrityDesign}.
\subsection{Tensegrity aerial vehicle}
This paper introduces tensegrity aerial vehicles, collision-resilient quadcopters designed with stiff icosahedron tensegrity structures. These vehicles incorporate the collision resilience of tensegrities and the mobility of quadcopters. To guide the design process of these vehicles, we propose a model-based approach that employs dynamics simulation to predict structural stresses during collisions, and to help us select components that can endure these stresses. Additionally, we create an autonomous re-orientation strategy to help the vehicles take off again after collisions. Exploiting the sphere-like geometry of the icosahedron, the tensegrity aerial vehicles can rotate from an arbitrary orientation on the ground to ones easy for takeoff. With collision resilience and re-orientation ability, tensegrity aerial vehicles can operate in cluttered environments without complex collision-avoidance strategies. Moreover, we further extend the vehicles' ability by adopting the inertial navigation method in \cite{wu2020using}, which enables the vehicles to perform short-range autonomous operations without external sensing. The resulting vehicles can thus serve as field robots and work on challenging tasks such as traversing through a cluttered corridor filled with smoke to search for survivors.

This work builds upon our previous study \cite{zha2020collision}, with following extensions: 1) A new dynamics simulation is introduced for guiding tensegrity structure design, accompanied by analysis to demonstrate the structural benefits of tensegrities. 2) A new stiff tensegrity shell design is presented. The rods of the shell are not rigidly connected, leading to enhanced collision resilience. 3) A new re-orientation strategy is proposed. The strategy can systematically determine the feasibility of rotations between tensegrity faces and calculate thrust commands from a desired torque using an optimization-based converter, fully utilizing the vehicle's thrust authority.

The contribution of this work are:
1) It presents a model-based approach for designing collision-resilient tensegrity aerial vehicles, supported by a dynamics simulation tool we have open sourced.
2) It proposes a re-orientation controller to facilitate flight resumption post-collision, and makes the corresponding development and analysis tools open sourced.
3) It validates the design approach and the controller with an experimental vehicle (Fig. \ref{VehiclePic}) and demonstrates its ability to survive collisions, resume flight, and perform autonomous operations in an unknown environment.
The source code of the dynamics simulation, the structural advantage analysis, and the re-orientation analysis is available at: \href{https://github.com/muellerlab/TensegrityAerialVehicle}{\tt github.com/muellerlab/TensegrityAerialVeh-\\icle}
\section{Design of the tensegrity shell}
\label{sec:tensegrityDesign}
In this section, we motivate the idea of protecting a quadcopter with a stiff icosahedron tensegrity shell, introduce the approach used to design the tensegrity with stress analysis based on a dynamics simulation, and showcase the structural advantage of the tensegrity design.

We choose to design the tensegrity aerial vehicle in the form of a quadcopter because the abilities to hover and to vertically take off and land make operations easier in cluttered environments. Meanwhile, we choose to protect the quadcopter with a 6-rod orthogonal icosahedron \cite{jessen1967orthogonal} tensegrity, whose near-spherical shape offers omnidirectional protection with minimal structural weight.

The tensegrity shell's primary role is to shield the quadcopter from damage during collisions. Hence, it must withstand impacts without breaking, and its deformation should be small to prevent external obstacles from making contact with the internal components. Consequently, the successful design of the tensegrity depends on the selection of components (rods and strings) that possess suitable stiffness and strength. 

We favor stiff components with little flexibility for two reasons. First, a stiff shell exhibits little deformation during collisions, requiring less buffer space to protect internal components like propellers from exposure. Thus, the tensegrity can be smaller in size, and this helps the vehicle to fit through narrow gaps. Second,  a stiff tensegrity reduces disruptive system vibrations, resulting in less flight disturbance. Meanwhile, we prefer lightweight components as they help retain the agility and flight time of the aerial vehicle. In the following subsection, we detail the process of determining whether certain components meet these design requirements.

\subsection{Stress analysis with dynamics simulation}
\begin{figure}[b]
    \centering
    \includegraphics[width=0.9\linewidth]{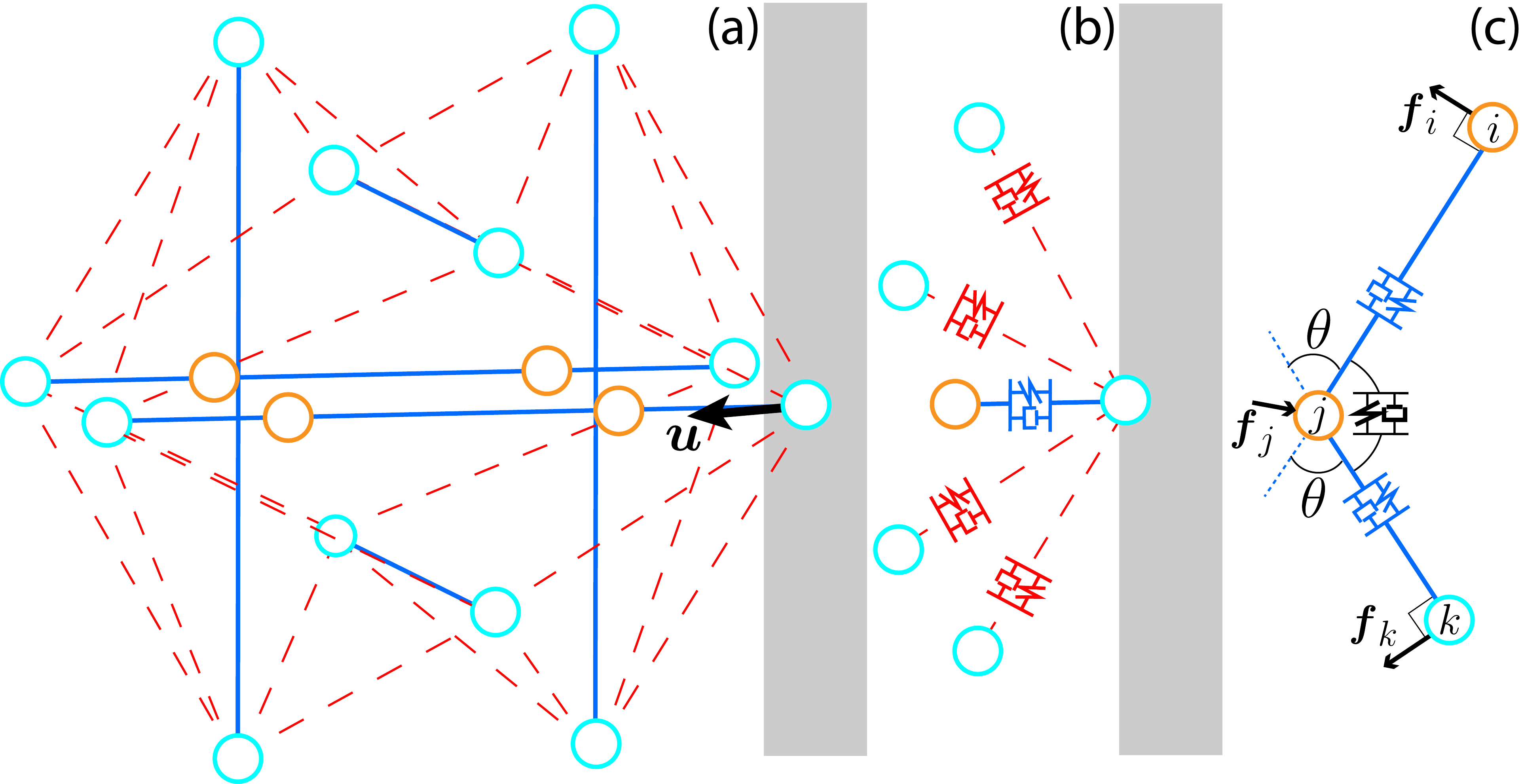}
    \caption{(a) The tensegrity vehicle is simplified as point masses in a stress network. Cyan spheres represent tensegrity nodes whereas orange spheres represent quadcopter nodes. (b) Strings and rods are modeled as massless spring-damper pairs. (c) Connections between two short rods are modeled as torsional spring-damper pairs.}
    \label{TensegrityModel}
\end{figure}

To predict if a tensegrity aerial vehicle can survive a collision, we simulate the dynamics of the tensegrity structure during the collision and calculate the stress in the structure with the simulation result. In contrast to the static stress analysis method we previously proposed in \cite{zha2020collision}, this dynamics method has two advantages. First, it accounts for the tensegrity deformation and captures the transient effects during the propagation of stress. Second, it considers the stress concentration caused by the mass of the quadcopter mounted on the tensegrity rods. These advantages lead to a more accurate stress estimate and allow us to easily verify if the tensegrity design meets the deformation criteria.

The tensegrity vehicle is modeled as point masses suspended in a stress network, as depicted in Fig. \ref{TensegrityModel}a. We define a \emph{tensegrity node} as the point where a rod connects to strings. An icosahedron tensegrity has 12 nodes, each is a point mass representing the mass of the fasteners at the node's position, as well as half of the rod and the strings connected to that node. Each tensegrity node connects to a rod and four strings, represented as a massless linear spring-damper pair (Fig. \ref{TensegrityModel}b).

The quadcopter is modeled using four evenly-distributed \emph{quadcopter nodes}, which are mass nodes attached to a pair of parallel rods in the tensegrity shell. Each quadcopter node represents the mass of propeller and motor at its location, and a quarter of the batteries and electronics. These quadcopter nodes divide the \emph{full-length rod} they're mounted on into three \emph{short rods}. Each connection between short rods is represented as a torsional spring-damper pair, as shown in Fig. \ref{TensegrityModel}c, with the torsional spring constant derived from the rod bending model, as we will show later. It is important to note that our model does not treat the rod hosting quadcopter nodes as a single, inflexible entity. Instead, it allows for relative rotation between short rods, facilitating the capture of transient stress concentration effects resulting from the uneven mass distribution caused by the mounting of the quadcopter.

We denote the nodes in the system as ${n}_i$, where $i = 1,...,12$ for the tensegrity nodes and $i = 13,...,16$ for the quadcopter nodes. The position of the $i^{th}$ node is represented by $\boldsymbol{x}_i$. For simplicity and consistency, variables related to rods are denoted with a superscript $r$ and those related to strings with a superscript $s$. The connectivity of the nodes can then be represented with indicator variables $N^r_{i,j}$ and $N^s_{i,j}$:
\begin{flalign}
 \label{connectivity_rod}
 & N^r_{i,j}=   \begin{cases}
                                  1 \mbox{, if a rod connects ${n}_i$ and ${n}_j$}\\ 0 \mbox{, otherwise} \end{cases}\\
 \label{connectivity_string}
 & N^s_{i,j}=   \begin{cases}
                                  1 \mbox{, if a string connects ${n}_i$ and ${n}_j$} \\
                                  0 \mbox{, otherwise} \\
  \end{cases}
\end{flalign}
We further define $T_{i,j}$ and $C_{i,j}$ as the value of the tensile force in string and compression force in rod connecting node ${n}_i$ and ${n}_j$. They can be calculated from Hooke's law, with a special modification that compressed strings generate no force: 
\begin{flalign}
\label{tensile force}
& T_{i,j} = \begin{cases}
N^s_{i,j}K^s(L_{i,j} - L^s) \mbox{, if $L_{i,j} \geq L^s$}\\
0 \mbox{, otherwise} \end{cases}\\
\label{compression force}
& C_{i,j} = N^r_{i,j}K^r_{i,j}(L^r_{i,j} -L_{i,j})
\end{flalign}
where $K^s$ and $K^r_{i,j}$ respectively are the spring constants of the string and the rod. We use subscripts $i$ and $j$ to specify rod-related variables, as these indices correspond to the two end nodes of the rod. $L_{i,j} = ||\boldsymbol{x}_i-\boldsymbol{x}_j||$ is the distance between node ${n}_i$ and ${n}_j$. $L^s$ and $L^r_{i,j}$ are the corresponding pre-deformation length of the string and the rod. Note that due to the existence of possible self-stress (also known as pre-tension) in the icosahedron tensegrity \cite{pellegrino1986matrix}, tensegrity components may be deformed even without an external load.

In addition to the tensile and compression forces, there exist linear damping forces which inhibit relative linear motion between nodes. We assume that each damping force is aligned with the corresponding string or rod, and its value is proportional to the relative velocity of the nodes: 
\begin{align}
{D}_{i,j} = (N^s_{i,j} + N^r_{i,j})\, c_{i,j}(\boldsymbol{\dot{x}}_j - \boldsymbol{\dot{x}}_i)^T \boldsymbol{e}_{i,j}
\end{align}
where $\boldsymbol{e}_{i,j}  \in \mathbb{R}^{3}$ is the unit vector pointing from ${n}_i$ to ${n}_j$ and $c_{i,j}$ is the corresponding linear damping coefficient.

Additionally, for the connections between short rods, which are modeled as torsional spring-damper pairs and shown in Fig. \ref{TensegrityModel}c, we assume the spring moment is proportional to the angle between the neighboring rod and the damping moment is proportional to the angular velocity:
\begin{align}
{M}_{j} = 
\begin{cases}
\xi_{j}\theta + c_j\dot{\theta} \mbox{, if $n_j$ connects two short rods}\\
0 \mbox{, otherwise} \end{cases}
\label{bendingMoment}
\end{align}
where $\xi_{j}$ is the torsional spring constant at $n_j$, derived from rod bending model, while $c_j$ is the torsional damping constant at $n_j$. For pure bending of a rod, the radius of curvature equals the product of Young's modulus $E^r$ and second moment of area $I^r$, divided by the bending moment ${M}_{j}$ \cite{beer2012mechanics}. Given that the bending angle is the ratio between the rod length and the radius of curvature, the torsional spring constant can be computed as:
\begin{align}
{\xi}_{j} = \frac{E^{r}I^{r}}{L_{i,k}}
\label{bendingStiffness}
\end{align}
As the mass of rods is assumed to be lumped at the nodes, the moment is equivalent to forces acting on nodes at the ends of the rods in orthogonal directions and a balancing force acting on the joint:
\begin{align}
\boldsymbol{f}_{i} =\frac{{M}_{j}}{L_{i,j}}\boldsymbol{e}_{\perp ji}, \quad \boldsymbol{f}_{k} =\frac{{M}_{j}}{L_{j,k}}\boldsymbol{e}_{\perp jk}, \quad \boldsymbol{f}_{j} = -(\boldsymbol{f}_{k} + \boldsymbol{f}_{i})
\label{bending_force_j}
\end{align}
where $\boldsymbol{e}_{\perp ji}$ and $\boldsymbol{e}_{\perp jk}$ are respectively unit vectors perpendicular to $\boldsymbol{e}_{j,i}$ and $\boldsymbol{e}_{j,k}$, pointing in the directions that would decrease the joint angle $\theta$.

Let $\boldsymbol{f}_{bi}$ represent the force due to bending on node $n_i$ and let $\boldsymbol{u}_{i}$ represent the external force acting on ${n}_i$. We use a method similar to \cite{goyal2018dynamics} and derive the equations of motion of the system with Newton's second law for each node $i$:
\begin{multline}
\boldsymbol{f}_{bi} + \boldsymbol{u}_{i}+ \sum_{j}(T_{i,j} -C_{i,j}+D_{i,j})\boldsymbol{e}_{i,j} = m_i\ddot{\boldsymbol{x}}_i
\label{ODE}
\end{multline} 
where $m_i$ is the mass of $n_i$.

To simulate the system's dynamics, we need to define the external forces that act on the tensegrity during the collision process. We estimate the force by simplifying the obstacle as a stiff linear spring, and assume the magnitude of the force acting on the tensegrity is proportional to the distance that the tensegrity node has penetrated the obstacle: 
\begin{align}
||\boldsymbol{u}_{i}|| = k_{o}p_{i}
\end{align}
where $k_{o}$ is the stiffness of the obstacle and $p_{i}$ is the penetration distance of node $n_i$. Studies on the stiffness of common obstacles like concrete walls can be found in the literature \cite{fenwick2000stiffness}. We assume the surface of the obstacle is frictionless, so reaction forces are normal to the surface of the obstacle.

We simulate the dynamics system described by Eq. \eqref{ODE} by providing the initial position and velocity of the nodes and numerically solving the corresponding initial value problem with the Radau method in the SciPy library \cite{SciPy2023}, which is chosen for its good general performance with stiff problems. 
The solution gives us positions of nodes over the simulated time, and we can then extract the tensile and compressive forces, as well as bending moment from these positions. 
The axial stress in the string or rod connecting nodes $n_i$ and $n_j$ can then be expressed as a function of simulation time as follows:
\begin{align}
\sigma^s_{i,j}(t) = \frac{T_{i,j}(t)}{A^s}, \quad \sigma^r_{i,j}(t) = \frac{C_{i,j}(t)}{A^r}  \label{StringStress}
\end{align}
where $\sigma^s_{i,j}$ and $\sigma^r_{i,j}$ are axial stress in corresponding strings and rods respectively. $A^s$ and $A^r$ are cross sectional areas of strings and rods. Furthermore, when considering the connection between two short rods, we can use the rod bending formula from \cite{beer2012mechanics} to compute the stress induced by bending at the rod's surface:
\begin{align}
\sigma^b_{j}(t) = \frac{M_{j}(t)r}{I^r}  \label{BendingStress}
\end{align}
where $r$ is the radius of rod. 
Thus, we can calculate the maximum stress at the node connecting two short rods as the sum of the bending stress at the rod surface and the maximum axial stress in the rods connected to it:
\begin{align}
\sigma^r_{j}(t) = \sigma^b_{j}(t) + \max_{i}(\sigma^r_{i,j}(t)) \label{NodeStress}
\end{align}
We can then use the computed stress information to check if the candidate components meet the design objectives. First, the stresses in the strings have to be smaller than their yielding strength $\sigma^{sy}$, with a factor of safety for string $\eta^s$: 
\begin{align}
\forall{i,j,t} \quad \eta^s\sigma^s_{i,j}(t)  < \sigma^{sy}
\label{StringStressLimit}
\end{align}
Second, we need to ensure that the axial stress in each rod is less than its yield strength $\sigma^{ry}$ and critical buckling strength $\sigma^{rb}_{i,j}$, with a safety factor for the rod $\eta^r$:
\begin{align}
\forall{i,j,t} \quad \eta^r \sigma^r_{i,j}(t)  < \min(\sigma^{ry},\sigma^{rb}_{i,j})
\label{RodStressLimit}
\end{align}
% Here we assume the material of the rod has symmetrical structural property so its yielding strength is the same for both compression and tension.
Here the rod's critical buckling strength can be approximated with Euler's buckling theory:
\begin{align}
\sigma^{rb}_{i,j} = \frac{\pi^2E^{r}I^{r}}{A^r(L^r_{i,j})^2}
\label{RodBucklingLimit}
\end{align}
Third, the stresses at the nodes connecting two short rods should also be smaller than the rod yielding strength with the safety factor for rod: 
\begin{align}
\forall{j,t} \quad \eta^r\sigma^r_{j}(t) < \sigma^{ry}
\label{RodJointStressLimit}
\end{align}

In addition to these stress conditions, we also need to ensure that the propellers and electronic components are not exposed during collisions. This can be done by computing the distances between the tensegrity surface and the quadcopter nodes, and ensuring that they are larger than a given threshold. By using this dynamics simulation along with stress checks and exposure checks, we can efficiently rule out candidate components that don't meet our design objectives without the need to physically construct and test the tensegrity structures.

\subsection{Structural advantage of the icosahedron tensegrity}
In an icosahedron tensegrity shell, external loads are dispersed among structural members as tension and compression, thereby avoiding large stress caused by bending. As a result, an icosahedron tensegrity shell can better survive collisions than common protective structures like propeller guards. We illustrate this structural advantage through a Monte Carlo study, which simulates wall-collision experiments and compares the maximum stresses in two aerial vehicle designs (a tensegrity and a propeller-guarded) during the collisions.
\begin{figure}[b]
    \centering
    \includegraphics[width=0.9\linewidth]{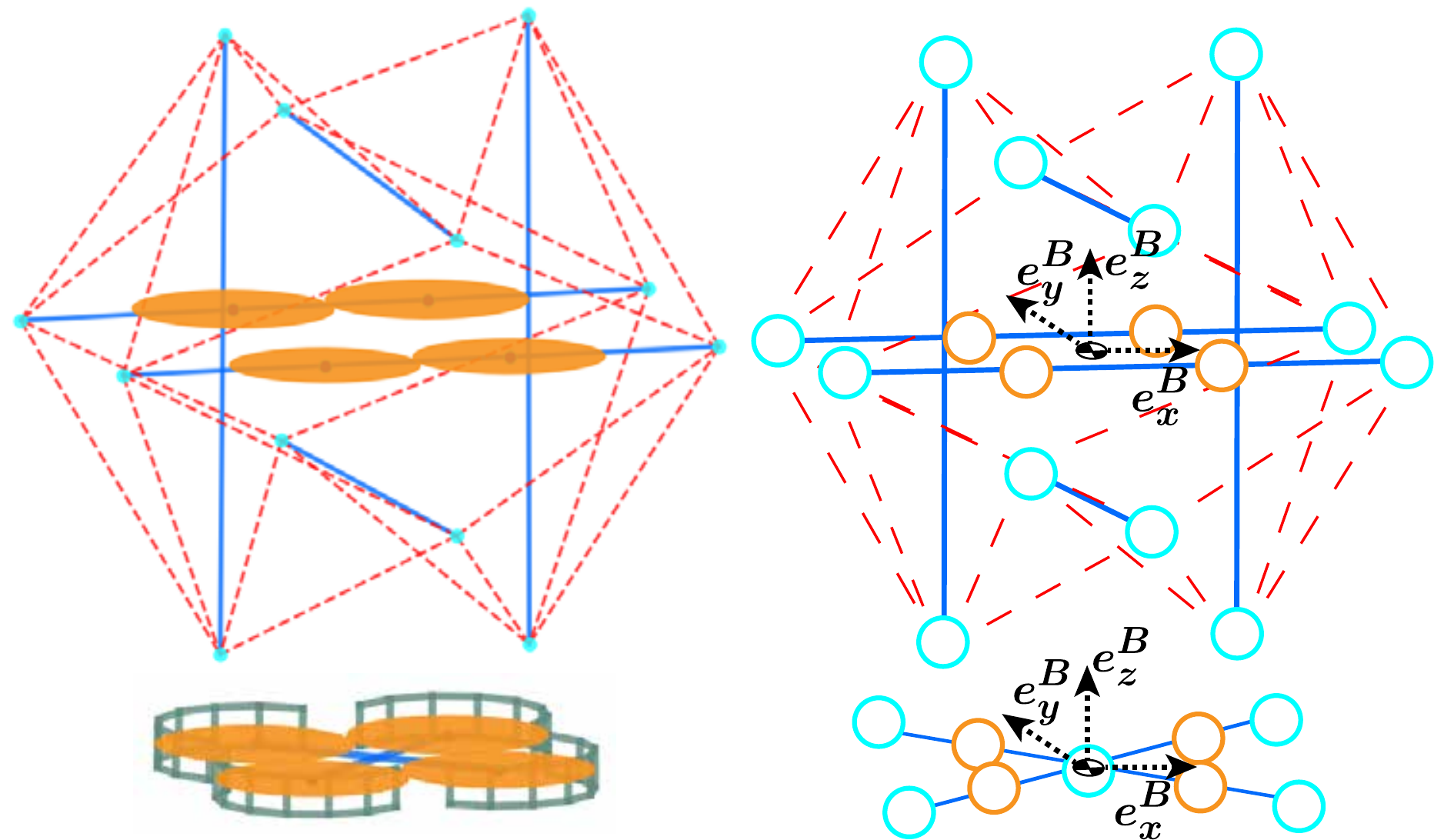}
    \caption{Left: illustration of the two collision-resilient aerial vehicles used for comparison. The top has a tensegrity shell whereas the bottom uses a propeller guard. Both vehicles have the smallest possible protection structure to host quadcopters with propellers of the same size. Right: we model both vehicles as point masses suspended in a stress network. We describe the vehicle's body-fixed frame with a set of three axes orthogonal to each other: $\boldsymbol{e}^B_x$, $\boldsymbol{e}^B_y$ and $\boldsymbol{e}^B_z$. Notice that for the tensegrity aerial vehicle, the quadcopter nodes are on the rods parallel to the $\boldsymbol{e}^B_x$ axis.}
    \label{StructureComparison}
\end{figure}

Both designs for our simulated experiments host a quadcopter with a total mass of $m_q$ and propellers of diameter $d$. The first design has the smallest tensegrity shell that can fully enclose the propellers, whereas the second design features the smallest propeller-guarded frame able to host the quadcopter. For simplicity, we depict the vehicles' body-fixed frames with three axes ($\boldsymbol{e^B}_x$, $\boldsymbol{e^B}_y$, $\boldsymbol{e^B}_z$) orthogonal to each other, as illustrated in Fig. \ref{StructureComparison}. We assume that the tensegrity shell and the propeller guard frame both have the same mass $m_s$ and are composed of solid cylindrical rods of identical material, thus sharing the same density $\rho^r$ and Young's modulus $E^r$. To fully define the tensegrity structure, we in addition specify $\gamma_m$, the ratio between the total mass of rods and total mass of strings in the tensegrity shell, and $F_s$, the pre-tension force in strings. It is worth noting that our analysis shows the maximum stress during a collision is insensitive to these parameters. Meanwhile, similar to the tensegrity model, the quadcopter with propeller guard is simplified as point masses in a network of rods modeled as massless linear spring-damper pairs connected by joints modeled as torsional springs and dampers. Notice that there is a minor difference in the model: for the joints connecting perpendicular rods, the rest angles corresponding to zero moment are $\frac{\pi}{2}$. Moreover, for both the tensegrity vehicle and the propeller-guard vehicle, we assume the dampers will make the corresponding systems, including the spring-damper pair and their directly-connected nodes, critically damped.
\color{black}
In our simulated experiments, we consider a wall with stiffness $k_o$. Before the collision, the tensegrity aerial vehicle and the propeller guard vehicle move perpendicularly toward the wall with a speed $v$. Both vehicles do not rotate before collisions. In the Monte Carlo study, we simulate 2000 experiments, each with a different random collision orientation generated by the following steps. First, we randomly sample points with a uniform distribution on the surface of a unit sphere attached to the vehicle's body-fixed frame. Then, we compute the collision orientation as the rotation which maps the vector pointing from the origin to the sampled point in the body-fixed frame to a vector pointing perpendicularly to the wall in an inertial frame attached to the wall. Due to the symmetry of both vehicles, we only need to study orientations corresponding to nodes sampled in a single octant (one of the eight divisions of the Euclidean space separated by the three orthogonal axes) of the sphere surface. 

The Monte Carlo study are conducted with key parameters in Table \ref{table:1}. The parameters correspond to a tensegrity shell made with carbon fiber rods and braided nylon string. Additionally, we choose $F_s=20$N, which corresponds to a stiff shell without large pre-tension stress in the system. 
Notice that given the same structural mass budget, the rods used in the tensegrity are longer and therefore thinner.
\begin{table}[htbp]
\centering
\caption{Key parameters used in the comparison simulation example}
\label{table:1}
\begin{tabular}{c|l}
\vspace{2pt}
Parameter   & Value \\ \hline
total structure mass & $m_s$ = 50g \\
total quadcopter mass & $m_q$ = 250g \\
string pre-tension & $F^s$ = 20N \\
rod-string mass ratio & $\gamma_m$ = 20 \\
rod density & $\rho^r$ = 2000$\text{kg/m}^3$ \\
string density & $\rho^s$ = 1150$\text{kg/m}^3$ \\
rod Young's modulus & $E^r$ = 3.2$\times10^{10}$Pa \\
string Young's modulus & $E^s$ = 4.1$\times10^{9}$Pa \\
diameter of 2.5-inch propellers & $d$ = $63$mm \\ 
wall stiffness & $k_o$ = 4.7$\times10^{7}$N/m\\ 
initial speed before collision & $v$ = $5$m/s \\ 
\end{tabular}
\end{table}

\begin{figure}
    \centering
    \includegraphics[width=\linewidth]{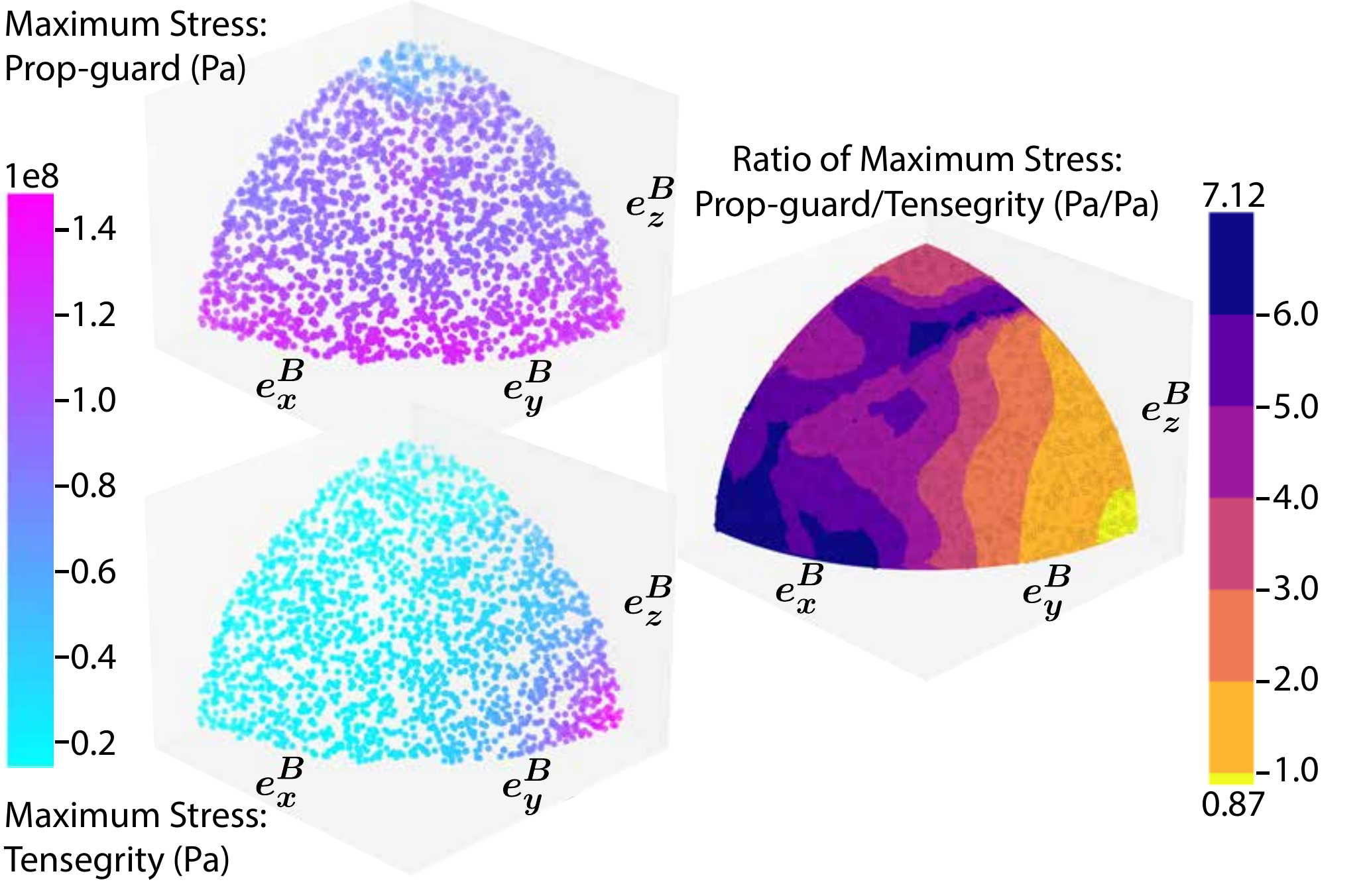}
    \caption{Visualization of the Monte Carlo study result. The positions of the points correspond to the collision orientation. Top left: scatter plot of the maximum stress in propeller guard during collisions. Bottom left: scatter plot of the maximum stress in tensegrity. Right: the ratio of the maximum stress in propeller guard to that in tensegrity. Larger values indicate more tensegrity advantage. The color on the surface is interpolated from the scattered simulated experiment data points.}
    \label{MaxStressInSystem}
\end{figure}

The result of the Monte Carlo study, visualized in Fig. \ref{MaxStressInSystem}, shows that the tensegrity holds a structural advantage over the propeller guard for collision resilience. Among the 2000 simulations, the tensegrity's mean maximum stress is $34.4$MPa, compared to $100.5$MPa in the propeller guard. For $80\%$ of the samples, the maximum stress in the tensegrity vehicle is smaller than half of that in the propeller-guarded vehicle. 
On the other hand, the propeller-guarded vehicle experiences a smaller maximum stress than the tensegrity aerial vehicle (i.e. propeller guard is superior) in only $2.7\%$ of the cases. Moreover, note that the high-stress points in the tensegrity plot are not symmetrically distributed. This comes from the non-uniform placement of quadcopter nodes, which are solely attached to the rods parallel to the $\boldsymbol{e^B}_x$ axis. Hence, the most severe stress is experienced when these rods collide perpendicularly with the wall. In such circumstances, the deformation within the tensegrity structure is restricted, leading to a less effective load distribution. This analysis result suggests that during high-speed operations, tensegrity aerial vehicles should avoid flying with these rods pointing forward to avoid structural failures.

\begin{figure}
\centering
    \includegraphics[width=\linewidth]{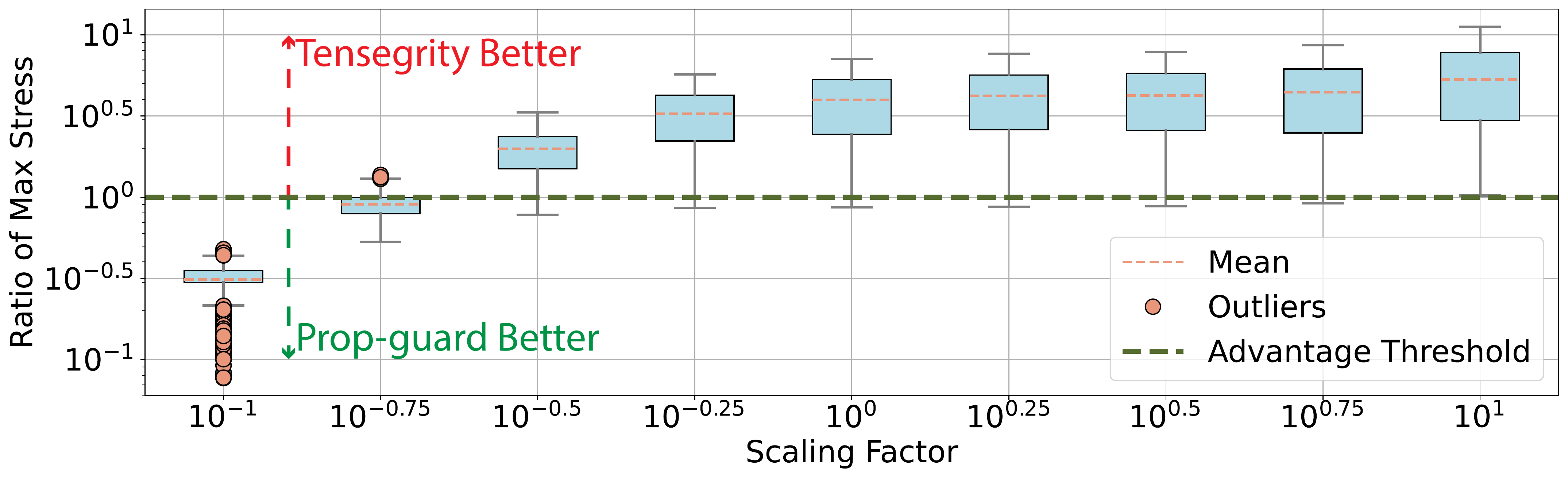}
    \caption{This figure illustrates the structural advantage of tensegrity over propeller guard for aerial vehicles of varying scales. The horizontal axis represents the scaling factor, while the vertical axis indicates the relative advantage, measured as the ratio of maximum stress endured by propeller guard to tensegrity during collision simulations. The rising trend suggests the tensegrity's advantage becoming more prominent with larger vehicle sizes.}
    \label{ScaleAnalysis}
\end{figure}

\color{black}   
In addition, we have investigated the structural advantage of the icosahedron tensegrity for vehicles of various scales. 
Specifically, we scale all length-related parameters linearly, all mass and force-related parameters cubically, while maintaining the ratios and material characteristics parameters from Table \ref{table:1}. We then conduct the same Monte Carlo analysis for the scaled tensegrity aerial vehicle and propeller-guarded vehicle and record the ratio of maximum stresses. 
The result of the analysis is shown in  Fig. \ref{ScaleAnalysis}.
As the vehicle size increases, the icosahedron tensegrity's collision resilience relative to the propeller guard also improves. As the propeller guard increases in size, bending becomes the primary source of stress due to the increased moment arm length. Consequently, as the vehicle scales up, the maximum stress in the tensegrity shell increases at a slower rate than that in the propeller guard. Conversely, as the vehicle size decreases, the propeller guard becomes more effective compared to the tensegrity. However, at smaller scales, the maximum stresses during collisions also decrease, and factors like air resistance become dominant, which in turn reduces the necessity for high-speed collision resilience. \color{black}

\section{Dynamics model and control of tensegrity aerial vehicles}
\label{sec:modellingAndControl}
In this section, we introduce the models and controllers of the tensegrity aerial vehicles. The vehicles primarily execute two types of motion: in-flight, they operate like standard quadcopters with a flight controller; on the ground, they employ a re-orientation controller to rotate themselves to an orientation with propellers pointing upward, preparing for takeoff.

Our previous research \cite{zha2020collision} proposed a strategy that repurposed the flight attitude controller for re-orientation. It relied on physical experiments to determine the feasibility of rotations and imposed a constraint of sum of thrusts being zero, limiting the rotational torque the vehicle could produce. This section introduces a new re-orientation strategy, which offers two improvements. First, it systematically determines rotation feasibility and plans re-orientation paths. Second, it incorporates a new thrust converter that optimizes the vehicle's re-orientation torque command while considering the thrust constraints. As a result, it increases the reliability of the re-orientation process.

This section aims to provide a generalized strategy for modeling and controlling tensegrity aerial vehicles. For discussions on the specific experimental vehicle we created, please refer to Section \ref{sec:experiment}. The code for related analysis is available in our open source repository (see link in Section \ref{sec:introduction}). 
\subsection{Vehicle dynamics and controller during flight}
Given the stiff shells of the tensegrity aerial vehicles, we make the assumption that the vehicles behave as rigid bodies when not in collision. As a result, the vehicles are modeled identically to standard quadcopters, and conventional quadcopter controllers are utilized for flight operations. % For more detailed information on quadcopter flight modeling and control, please refer to \cite{koubaa2021unmanned}.\color{black}
\subsection{Vehicle dynamics for re-orientation}

\begin{figure}
    \centering
    \includegraphics[width=\linewidth]{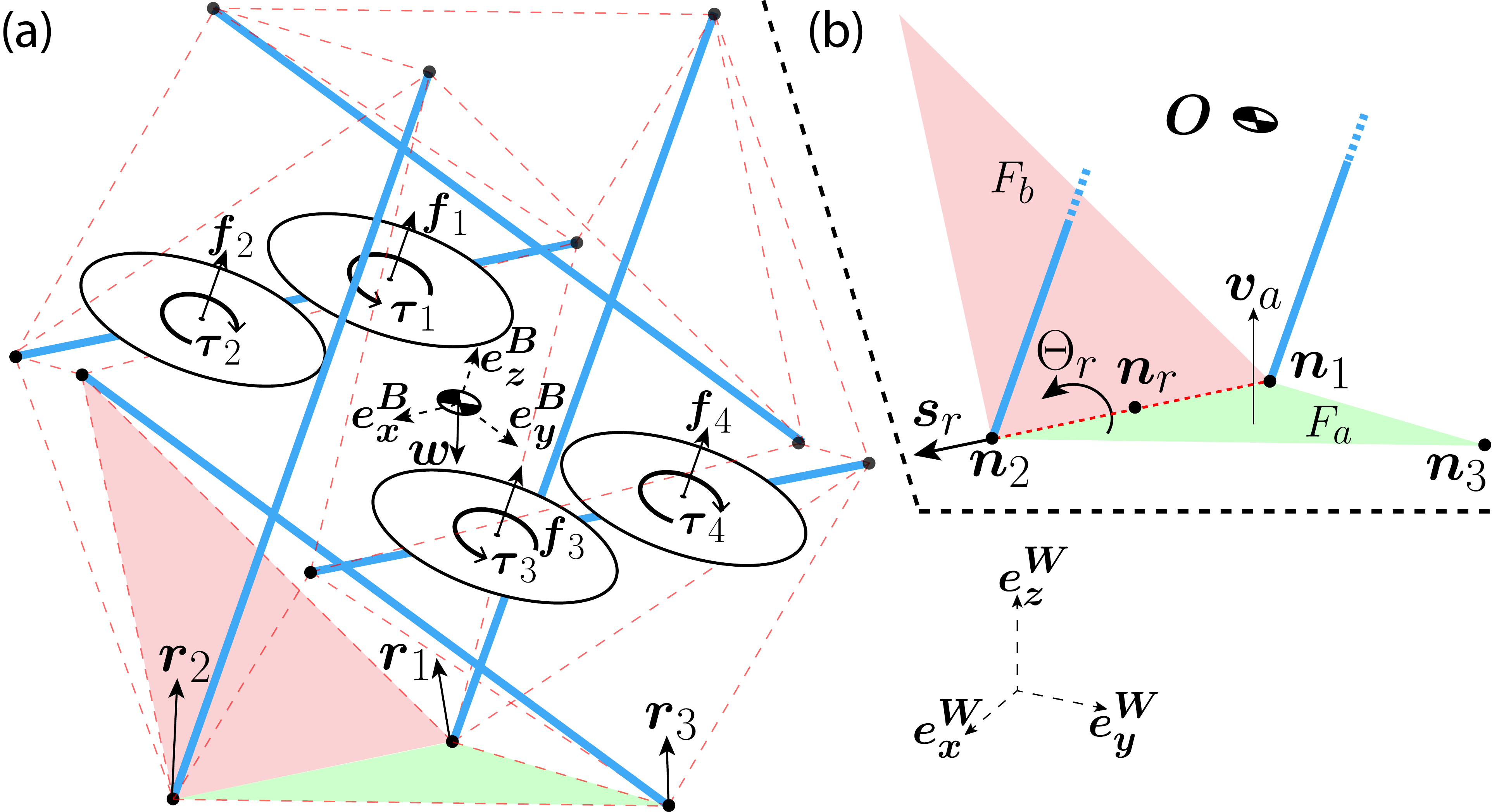}
    \caption{The figure shows the dynamics model of a tensegrity aerial vehicle during a rotation between faces. (a): The vehicle experiences weight $\boldsymbol{w}$, thrusts and yaw torques generated by the propellers $f_i$ and $\tau_i$ (where $i=1,2,3,4$), and contact forces $\boldsymbol{r}_j$, where $j= 1,2,3,\ldots$. $\boldsymbol{W}$ denotes the world frame fixed to the ground and $\boldsymbol{B}$ denotes the vehicle's body-fixed frame. (b): The tensegrity rotates from face $F_a$ to its neighbor face $F_b$. We study the rotation about rotation point $\boldsymbol{n}_r$ with a rotation axis $\boldsymbol{s}_r$ and a total rotation angle $\Theta_r$. The tensegrity contacts the environment at multiple points denoted by $\boldsymbol{n}_j$.}
    \label{quadModel}
\end{figure}
The model used to describe the dynamics of a tensegrity aerial vehicle during re-orientation is shown in Fig. \ref{quadModel}a. External forces and torques include the weight of the vehicle, thrust and torque from propellers, and reaction forces and torques due to contact with the environment.

The vehicle's attitude is defined as a rotation matrix $\boldsymbol{R}$, mapping vectors from the body-fixed frame $\boldsymbol{B}$, which is affixed to the center of mass $\boldsymbol{o}$, to the world frame $\boldsymbol{W}$, which is inertial and affixed to the ground, i.e., $\boldsymbol{v^W} = \boldsymbol{R}\boldsymbol{v^B}$. To avoid possible confusion, when a vector is used in analysis across different frames, we use superscript to indicate in which frame the vector is expressed.

The translational dynamics comes from Newton's law:
\begin{align}
m\boldsymbol{\ddot{d}} = \boldsymbol{w^W}+\boldsymbol{R}\boldsymbol{e^B}_z \sum_{i=1}^4f_i + \sum_j\boldsymbol{r^W}_j
\label{TranslationalMotion}
\end{align}
where $m$ is the vehicle mass, $\boldsymbol{w}$ is its weight, $\boldsymbol{d}$ is its position relative to a fixed point in the world frame, $f_i$ is the thrust generated by the $i^{th}$ propeller, $\boldsymbol{e^B}_z$ is a unit vector pointing along the $z$-axis of the body-fixed frame, and $\boldsymbol{r}_j$ represents the reaction force from the environment acting on node $j$.

Rotational dynamics is modeled with Euler's equation:
\begin{equation}
\begin{split}
&\boldsymbol{J}\dot{\boldsymbol{\omega}} + \skewSym{\boldsymbol{\omega}}\boldsymbol{J}\boldsymbol{\omega} =
\sum_{i=1}^4f_i\boldsymbol{m}_{o,i} + \sum_j(\skewSym{\boldsymbol{n^B}_j}\boldsymbol{r^B}_j)
\end{split}
\end{equation}
where $\skewSym{\cdot}$ maps a $\mathbb{R}^{3}$ vector to a corresponding $\mathbb{R}^{3 \times 3}$ skew-symmetric matrix. Left multiplying $\skewSym{\cdot}$ is equivalent to the cross product. $\boldsymbol{J}$ is the moment of inertia tensor of the vehicle with respect to its center of mass, and $\boldsymbol{n}_j$ is the position of the node $j$ that is in contact with the environment. Meanwhile, the angular velocity vector of the vehicle, $\boldsymbol{\omega} \in \mathbb{R}^{3}$, represents the rotation velocity between the body-fixed frame and the world frame, and it relates to the attitude matrix as follows:
\begin{equation}
\boldsymbol{\dot{R}} = \boldsymbol{R}\skewSym{\boldsymbol{\omega}}
\end{equation} 
Moreover, $\boldsymbol{m}_{o,i} \in \mathbb{R}^{3}$ represents the torque with respect to the center of mass generated by the $i^{th}$ propeller with a unit thrust, and can be computed as:
\begin{align}
\boldsymbol{m}_{o,i} = \skewSym{\boldsymbol{p^B}_i}\boldsymbol{e^B}_z + h_i\kappa\boldsymbol{e^B}_z,
\label{MixerMatrixCol_COM}
\end{align}
where $\boldsymbol{p}_i$ is the position of propeller $i$, $h_i$ denotes the handedness of the propeller $i$ ($1$ for right-handed, $-1$ for left-handed), and $\kappa$ is the propeller's torque coefficient. 
On the right of Eq. \eqref{MixerMatrixCol_COM}, the first term represents the torque coming from the cross product of the moment arm and its respective thrust force, whereas the second term shows the drag torque from propeller rotation.
\subsection{Re-orientation strategy}
\begin{figure}[b]
    \centering
    \includegraphics[width=0.7\linewidth]{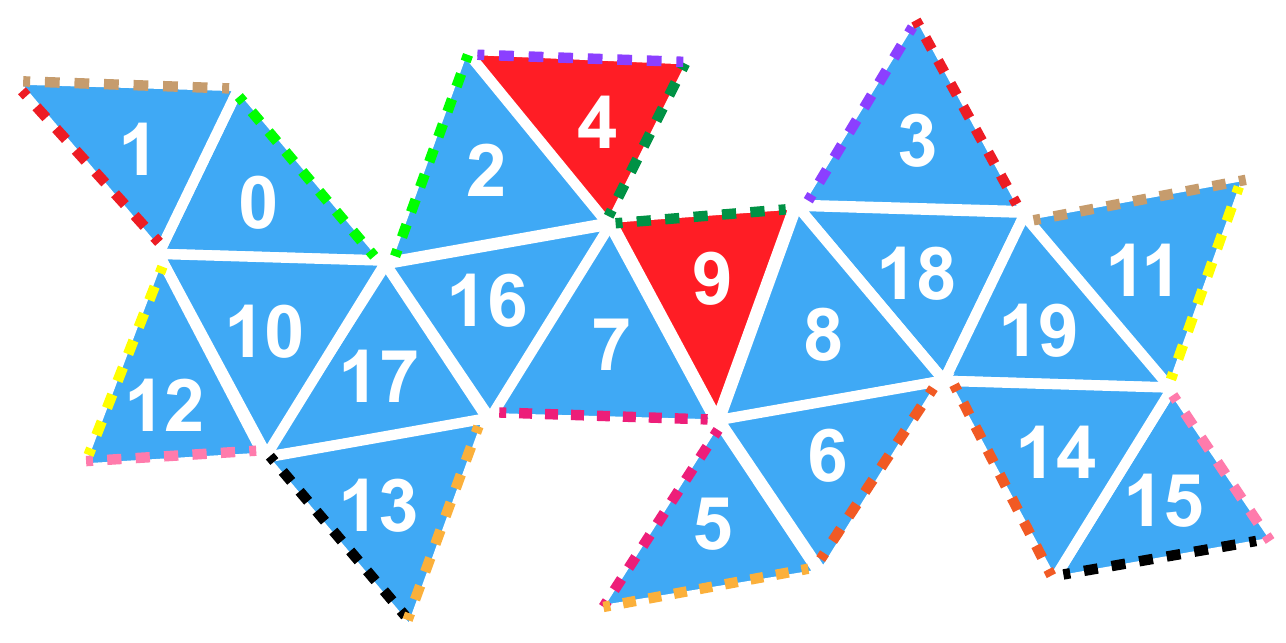}
    \caption{Faces of an unfolded icosahedron tensegrity. The tensegrity can take off when face 4 or 9 is contacting the ground. Dashed lines with the same color indicate overlapped edges when the tensegrity is folded back.}
    \label{TensegrityFace}
\end{figure}

To facilitate the resumption of flight after collisions, a re-orientation controller is created to rotate the vehicles from arbitrary orientations to ones easy for takeoff. An icosahedron tensegrity has twenty faces, and we define two faces as neighboring if they share two nodes. Assuming the tensegrity is on flat ground, we denote the face in contact as $F_i$ if the $i^{th}$ face is touching the ground.  Fig. \ref{TensegrityFace} shows an unfolded icosahedron, illustrating the neighboring relationship of tensegrity faces. When face 4 or 9 (highlighted in the figure) is the contact face, the propellers point upward, indicating the tensegrity is prepared for takeoff. Thus, the objective of the re-orientation is to rotate the tensegrity aerial vehicle so that face 4 or 9 becomes the contact face.

The re-orientation strategy decomposes the task into a sequence of rotations between neighboring tensegrity faces, offering two key benefits. First, each rotation is simple to model, as the rotation axis corresponds to the line shared by the neighboring faces and the total rotation angle is determined by the icosahedron shape. Second, the strategy simplifies the problem into a finite state machine, thus enhancing robustness. If a rotation fails and the vehicle lands on an unexpected face, the controller can re-plan the path and continue with the task.

We define the re-orientation paths as a series of desired rotations to rotate the vehicle from the starting faces to a goal face. To find these paths, we follow a two-step procedure. First, we create a connection graph where the nodes represent the contact faces, and directed edges indicate feasible rotations between neighboring faces. Then, we search on this graph to find the shortest path from any starting face to the goal face.

The feasibility of a rotation between neighboring faces is evaluated by assessing whether the tensegrity aerial vehicle can generate a set of thrusts that counteract the gravitational torque, without causing the vehicle to slide or leave the ground. We assume that the electronic speed controllers (ESCs) of the vehicles are configured to drive the propellers bi-directionally, enabling the vehicles to generate additional torque for re-orientation. For the following analysis, we use the notation illustrated in Fig. \ref{quadModel}b, where $F_a$ denotes the starting face and $F_b$ represents the neighboring face to rotate to. We denote $\boldsymbol{n}_{r}$ as the rotation point, $\boldsymbol{s}_{r}$ as the rotation axis in the body-fixed frame, and $\Theta_{r}$ as the rotation angle. For a rotation to be feasible, there must exist a set of thrusts $[f_1,f_2,f_3,f_4]$ and reaction forces $[\boldsymbol{r}_1,\boldsymbol{r}_2]$ satisfying following conditions:
\begin{empheq}[left=\empheqlbrace]{alignat=2}
&\boldsymbol{e^B}_{z}\sum_{i=1}^4 f_i + \sum_{j=1}^2 \boldsymbol{r^B}_j + \boldsymbol{w^B} = \boldsymbol{0}
\label{force_eq_edge}\\
&\sum_{i=1}^4f_i\boldsymbol{m}_{r,i} + \sum_{j=1}^2 \boldsymbol{S}(\boldsymbol{n^B}_j-\boldsymbol{n^B}_r)\boldsymbol{r^B}_j  = \boldsymbol{S}(\boldsymbol{n^B}_{r})\boldsymbol{w^B}\label{moment_eq_edge}\\
&0 \leq \boldsymbol{r^B}_j \cdot \boldsymbol{v^B}_a, \forall j \in \{1,2\} \label{normalConstraint}\\
&||\boldsymbol{r^B}_j-(\boldsymbol{r^B}_j \cdot \boldsymbol{v^B}_a)\boldsymbol{v^B}_a||_{2} \leq \mu(\boldsymbol{r^B}_j \cdot \boldsymbol{v^B}_a), \forall j \in \{1,2\}  \label{fricConstraint}\\
&f_{min}\leq f_i \leq f_{max}, \forall i \in \{1,2,3,4\} \label{thrustrange}
\end{empheq}

The above conditions specify the scenario when the tensegrity vehicle fully compensates its gravitational torque and is about to initiate rotation. At this moment, the third contact point is about to leave the ground. Thus, its reaction force goes to zero and is therefore not included in the equations. Eq. \eqref{force_eq_edge}, derived from Newton's law, describes the force balance, while Eq. \eqref{moment_eq_edge} illustrates the balance of moments about the rotation point, $\boldsymbol{n}_r$. We use $\boldsymbol{m}_{r,i} \in \mathbb{R}^{3}$ to represent the torque with respect to $\boldsymbol{n}_r$ generated by a unit thrust of propeller $i$. Similar to Eq. \eqref{MixerMatrixCol_COM}, it is computed as:
\begin{align}
\boldsymbol{m}_{r,i} = \skewSym{\boldsymbol{p^B}_i-\boldsymbol{n^B}_{r}}\boldsymbol{e^B}_z + h_i
\kappa\boldsymbol{e^B}_z,
\label{MixerMatrixColumn_nr }
\end{align}
The constraint \eqref{normalConstraint} assures that the two nodes retain contact with the ground, as the reaction forces have non-negative components along $\boldsymbol{v}_a$, the unit ground-normal vector. Meanwhile, the no sliding condition is captured by \eqref{fricConstraint}, where $\mu$ denotes the friction coefficient between the vehicle and the ground. The feasible propeller thrust range is given in \eqref{thrustrange}, where $f_{max}$ represents the maximum thrust each propeller can generate, and $f_{min}$ symbolizes the negative thrust value produced when the propeller spins reversely at peak speed.

Based on the feasibility analysis, a connection graph can be constructed. A fully connected graph indicates that the vehicle can re-orient to any contact face. In contrast, the presence of disconnected nodes on the graph indicates that the vehicle is incapable of leaving the corresponding contact faces through rotation. This suggests that the vehicle fails to generate enough torque to counterbalance the gravitational torque under the thrust range, the contact constraint and no-sliding constraint. A design update incorporating stronger motors and/or longer moment arms, and a recheck of re-orientation feasibility are recommended to solve the problem. Once the connection graph has been constructed, the shortest paths from any starting face to the goal faces can be determined. Section \ref{sec:experiment}.B provides an example of generating the re-orientation paths for our experimental vehicle.
% Notice a reassessment of feasibility with the above process is required to ensure the effectiveness of the update.
\subsection{Reference rotation trajectory for re-orientation}
For each re-orientation step, the controller first identifies the rotation required for the face change, generates a reference rotation trajectory, and then tracks the generated trajectory. We employ a two-piece trajectory, which accelerates from a stationary state to the maximum angular velocity for the first half of the duration and then decelerates to stop for the second half. The angular acceleration remains constant in magnitude throughout, with a direction change at the midpoint of the duration. We have opted for this trajectory since it allows for straightforward tuning of the reference angular acceleration (which determines the aggressiveness of the trajectory) by adjusting the total trajectory time, denoted as $T$.
Note that the total rotation angle $\Theta_r$ is constant, so the magnitude of the reference angular acceleration solely depends on $T$:
\begin{align}
||\ddot{\boldsymbol{\Theta}}_{ref}|| = \frac{4\Theta_r}{T^2}
\label{AngAccValue}
\end{align}
Ideally, in the absence of tracking error, the total torque command equals the sum of the gravitational torque offset and the torque to generate the desired angular acceleration, which is inversely proportional to $T^2$. Thus, for large $T$, the offset predominantly dictates the total torque command. As $T$ decreases, tracking torque becomes dominant, and even a small adjustment in $T$ can significantly alter the torque command. Hence, when tuning $T$, we recommend beginning with a large initial value and then reducing it incrementally until the desired rotational behavior is achieved, all while ensuring the feasibility of the thrusts.

At a given time $t$, we can express the reference state with a reference rotation vector $\boldsymbol{{\Theta}}_{ref}$, a reference angular velocity vector $\boldsymbol{\dot{\Theta}}_{ref}$, and a reference angular acceleration vector $\boldsymbol{\ddot{\Theta}}_{ref}$. All three vectors point along the rotation axis $\boldsymbol{s}_{r}$. Moreover, we can find the reference vehicle attitude at $t$ from the reference rotation vector as:
\begin{align}
\boldsymbol{R}_{ref}(t) = \boldsymbol{R}_{s}f_{\boldsymbol{Rv}}(\boldsymbol{{\Theta}}_{ref}(t)),
\end{align}
where $\boldsymbol{R}_{s}$ is the attitude of the vehicle before the start of rotation, and $f_{\boldsymbol{Rv}}(\cdot)$ converts a rotation vector to its corresponding rotation matrix \cite{shuster1993survey}.
\subsection{Tracking controller for re-orientation}
To track the reference trajectory, we design a controller which generates a desired angular acceleration reducing the error as a second-order system:
\begin{align}
\boldsymbol{\ddot{\Theta}}_{d} = \boldsymbol{\ddot{\Theta}}_{ref} + 2\zeta_r \omega_r (\boldsymbol{\dot{\Theta}}_{ref} - \measAngVel) + \omega_r^2(\boldsymbol{\delta}_{r}) 
\end{align} 
where $\zeta_r$ is the desired damping ratio, $\omega_r$ is the desired natural frequency of the rotation and $\measAngVel$ is the angular velocity reading from the rate gyroscope. The attitude error, $\boldsymbol{\delta}_{r}$, represented as a rotation vector in the body-fixed frame, can be computed as:
\begin{align}
\mvec{\delta_{r}} =f_{\boldsymbol{vR}}(\boldsymbol{R}^{-1}\boldsymbol{R}_{ref})
\end{align}
where $f_{\boldsymbol{vR}}(\cdot)$ is the inverse of $f_{\boldsymbol{Rv}}(\cdot)$ and it converts a rotation matrix to a rotation vector.

The total desired torque command to track the trajectory can then be computed as the sum of the torque needed to offset gravity and the torque required to track the trajectory:
\begin{align}
\boldsymbol{\tau}_d = \boldsymbol{J}_r\boldsymbol{\ddot{\Theta}}_{d} + \skewSym{\hat{\boldsymbol{\omega}}}\boldsymbol{J}_r\hat{\boldsymbol{\omega}}-\boldsymbol{\tau}_g
\end{align}
where $\boldsymbol{J}_r$ is the mass moment of inertia of the tensegrity aerial vehicle with respect to the rotation point. $\boldsymbol{\tau}_g$ is the gravitational torque offset and can be computed as the cross product between the vector pointing from rotation point to the center of mass and the gravitational vector.

Next, we convert the torque command to per-propeller thrust commands that the vehicle can directly implement. The mapping from the thrusts to the generated torque is:
\begin{align}
\boldsymbol{\tau}_{p} = \sum_{i=1}^4 f_i\boldsymbol{m}_{r,i}
\label{ThrustTorque}
\end{align}

Notice torque $\boldsymbol{\tau}_{p} \in\mathbb{R}^{3}$. Hence, Eq. \eqref{ThrustTorque} forms a linear system with three equations and four unknown thrusts, leading to an under-determined mapping from torque to thrust. Moreover, due to the physical limits of the motors and the propellers, we also need to take thrust saturation into account. To find the thrust commands, we solve the following problems: 

When no thrusts are saturated and the exact desired torque can be generated, we command thrusts that minimize the sum of squares of the thrusts:
\begin{align}
\begin{aligned}
\min_{f_{i}}  & \sum_{i=1}^4 f^2_i\\
\textrm{s.t.  } &  \boldsymbol{\tau}_{d} = \sum_{i=1}^4 f_i\boldsymbol{m}_{r,i}\\
&  f_{min}  \leq f_i \leq f_{max}
\label{thrustMappingFeasible}
\end{aligned}
\end{align}

If problem \eqref{thrustMappingFeasible} has no feasible solution, force saturation is unavoidable. This usually happens when the controller tries to correct a larger-than-expected tracking error. In these cases, we solve for a set of feasible thrusts that minimize the norm of the error between the desired torque and the torque that can be generated, while taking into account the constraint of thrust generation authority:
\begin{align}
\begin{aligned}
\min_{f_{i}} & ||\sum_{i=1}^4 f_i\boldsymbol{m}_{r,i} - \boldsymbol{\tau}_d||_2\\
\textrm{s.t.  } 
&  f_{min}  \leq f_i \leq f_{max}
\label{thrustMappingInfeasible}
\end{aligned}
\end{align}

Both optimization problems \eqref{thrustMappingFeasible} and \eqref{thrustMappingInfeasible} are of relatively low dimension. Consequently, they can be solved in real-time, even on embedded systems with limited computation power, using tools like CVXGEN \cite{mattingley2012cvxgen}. A figure illustrating the thrust mapping for an example rotation problem and demonstrating the advantage of the optimization-based thrust converter introduced above can be found in Section \ref{sec:experiment}.B.

\section{Validation with experimental vehicle}
\label{sec:experiment}
In this section, we present the experimental tensegrity aerial vehicle, and the tests and analysis demonstrating its abilities.
\subsection{Experimental vehicle}
An experimental vehicle (see Fig. \ref{VehiclePic}) has been designed and built to validate the proposed vehicle functionalities. We have designed the tensegrity shell with rods of 20mm length, so the shell can fit a micro-scale quadcopter inside while still able to pass through narrow gaps between obstacles. Our design aims at providing protection against collisions at a target operation speed of 6m/s. We employ the design methodology described in Section \ref{sec:tensegrityDesign} to identify suitable candidate materials for constructing the tensegrity shell. Among all candidates that satisfy the design requirements, carbon fiber rods with 6mm outer diameters and braided nylon strings have been selected based on factors such as weight, cost, and availability. The vehicle weighs 300g. Each motor can generate a maximum thrust of 2.8N for a short time (for re-orientation) or 2.2N continuously (for flight). The thrust-to-weight ratio is about 3:1. The mass breakdown of the vehicle is as follows. 
\begin{center}
\begin{tabular}{| c | c | c | c |}
\hline
  Shell & Batteries & Electronics & Motors\\  \hline
  95g & 75g & 50g & 80g\\ 
\hline
\end{tabular}
\end{center}
\begin{figure}[b]
    \centering
    \includegraphics[width=\linewidth]{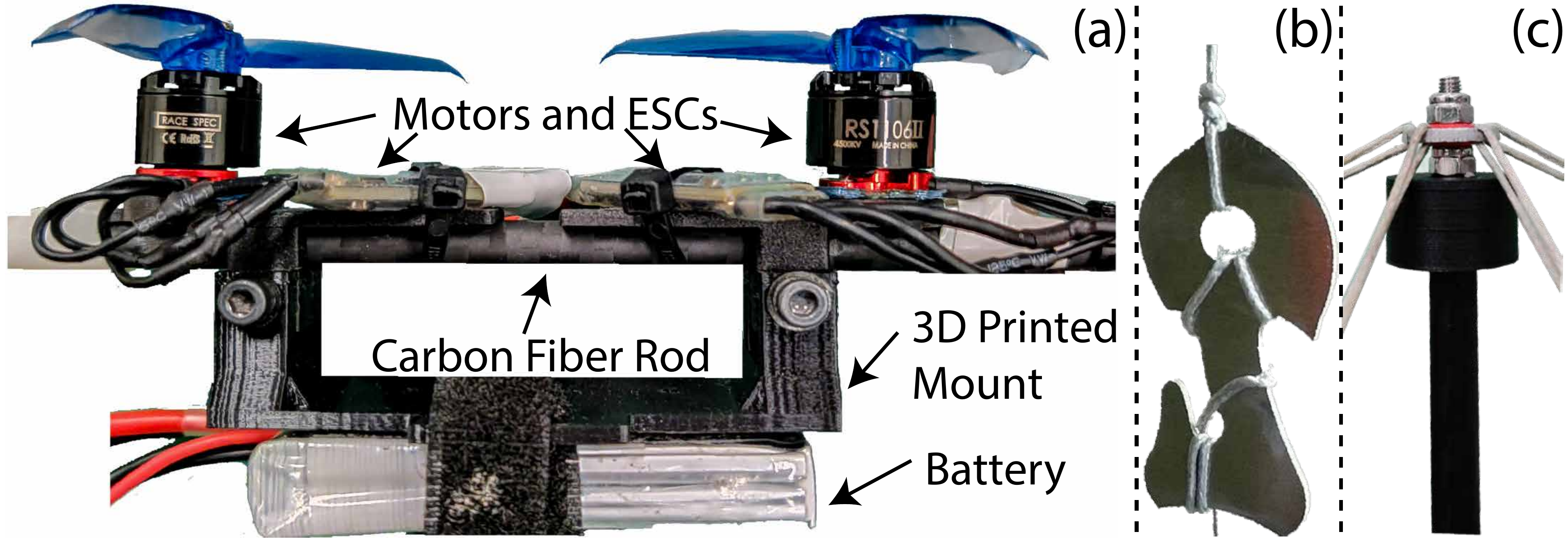}
    \caption{(a) Side-view of the tensegrity vehicle. Each horizontal rod has a battery and a pair of motors and ESCs attached to it. (b) Tension hook. (c) End cap connecting a rod to strings.}
    \label{TensegrityParts}
\end{figure}
% \color{blue}[R1.1] \color{purple} In comparison with its predecessor in \cite{zha2020collision}, the experimental vehicle retains an identical dimensions but weighs 50g more. The additional weight comes from the enhancements in the tensioning mechanism and connections between rods and strings. As a result, the tensegrity is easier to manufacture and more collision-resilient\color{black}.

The resulting vehicle does not have a flat frame commonly used in quadcopter designs. Instead, its motors and computation units are directly mounted to the tensegrity shell using custom-designed 3D-printed mounts.
Furthermore, to ensure even weight distribution, the design is powered by two batteries connected in series. Each battery is attached to one of the horizontal rods of the shell (see Fig. \ref{TensegrityParts}a). The design uses tension hooks to adjust self-stress in the tensegrity (see Fig. \ref{TensegrityParts}b) and 3D-printed end caps with fiber-glass infill to secure connections between rods and strings (see Fig. \ref{TensegrityParts}c).

\subsection{Collision resilience}
A drop experiment and an in-flight collision experiment were conducted to verify the collision resilience of the experimental vehicle. In the drop experiment, we dropped the vehicle to a concrete pavement to find the collision speed it can survive. The vehicle successfully survived a drop from a 7m tall balcony with a landing speed of 11.7m/s. When we drop the vehicle from the next available balcony of 10.5m with a landing speed of 14.4m/s, the tensegrity failed (a string snapped). For comparison, a 250g quadcopter built with a commercial propeller-guarded frame \cite{betafpv} hosting the same propellers and electronics as the experimental vehicle, fractured after a 3.25m drop with a 8.0m/s landing speed. 

\begin{figure}
    \centering
    \includegraphics[width=\linewidth]{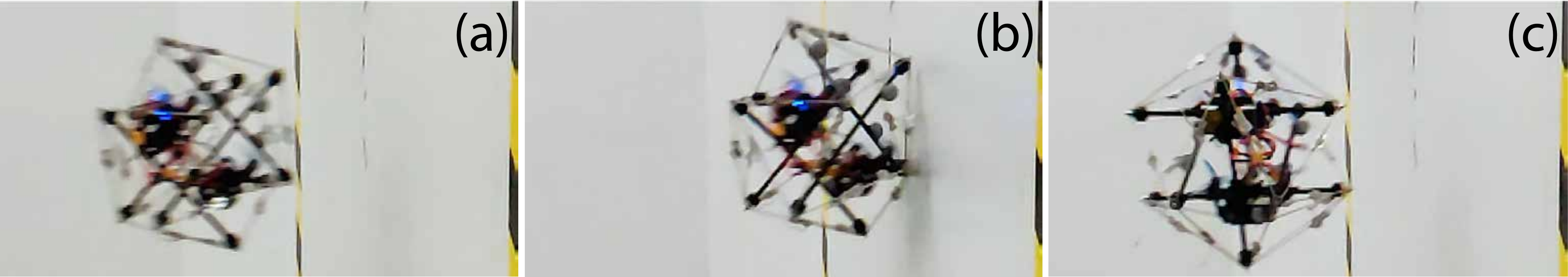}
    \caption{Sequence of images showing the process of a collision against a concrete wall: (a) Vehicle accelerates towards the wall. (b) Vehicle comes to a full stop. (c) Vehicle bounces back from the wall. The collision speed is 7.8m/s.}
    \label{Collision}
\end{figure}

In addition to the drop test, we also controlled the experimental vehicle to accelerate towards a concrete wall and collide with it to confirm the vehicle's ability to survive collision during an actual flight. An image sequence from a high-speed video of the collision process is displayed in Fig. \ref{Collision}. The vehicle survived a collision of 7.8m/s, the fastest flying speed it can reach under the space limit of our flight space. All components within the tensegrity structure remained intact throughout the process, and the vehicle retained its ability to fly post-collision. Videos of all collision experiments are included in the attached materials.

\subsection{Re-orientation}
\begin{figure}
    \centering
    \includegraphics[width=\linewidth]{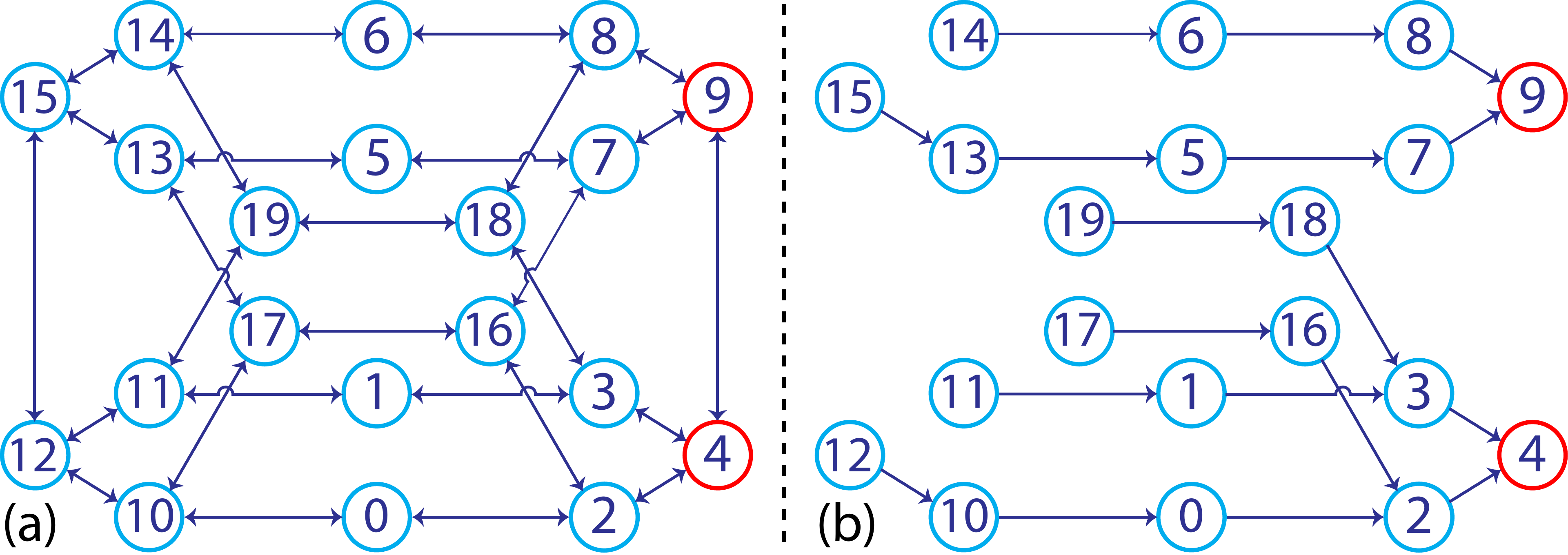}
    \caption{Generation of reorientation path. Nodes represent contact faces, with red denoting the goal faces to rotate towards. (a) A graph of all feasible face rotations is generated. Arrows indicate feasible rotations. (b) The shortest paths for each face to rotate to its closest goal face are generated, with arrows showing rotation directions.}
    \label{feasibleConnection}
\end{figure}

\begin{figure}
    \centering
    \includegraphics[width=\linewidth]{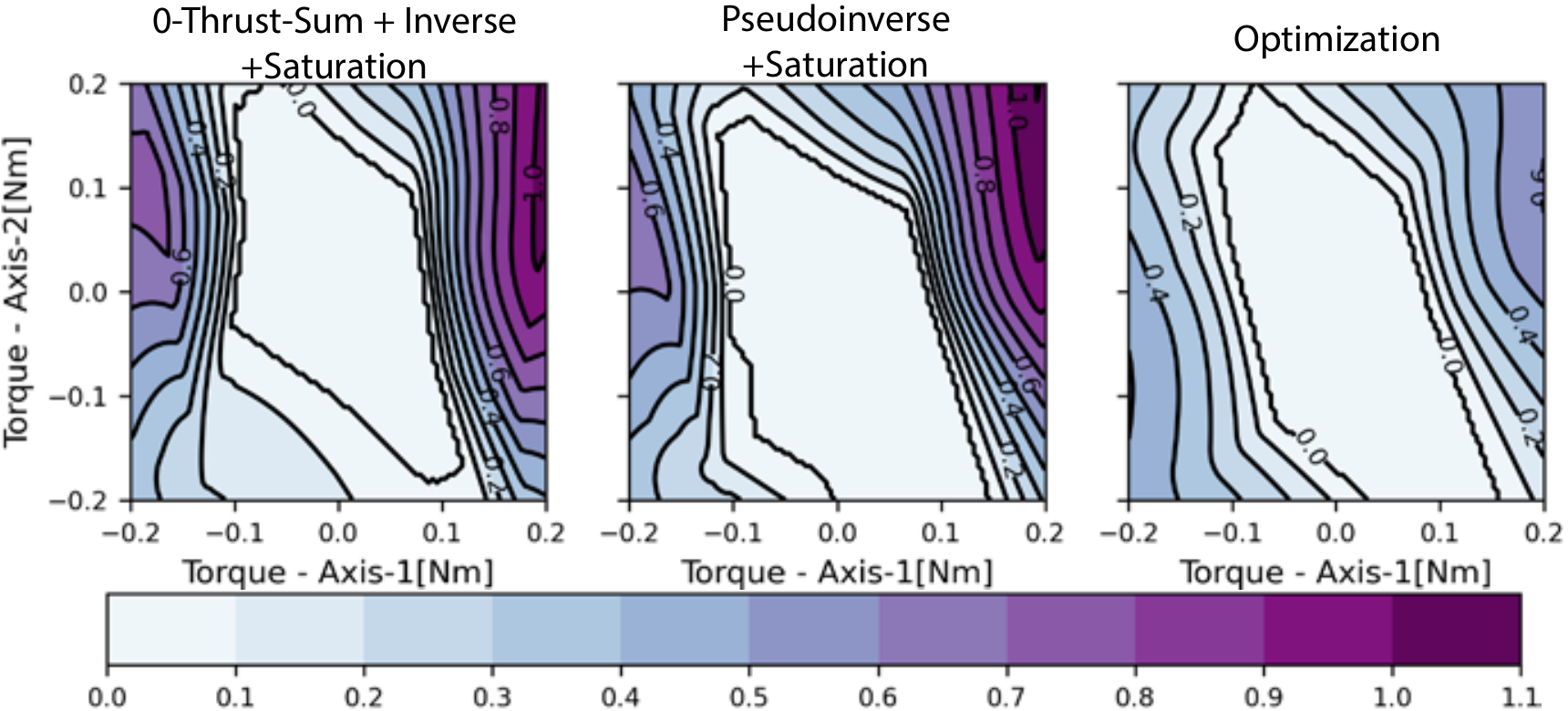}
    \caption{Thrust conversion error rates for rotation from face 3 to 4 with three different methods. Axis-1 is the desired rotation axis as shown in Fig. \ref{quadModel} and axis-2 points from the rotation point to the center of mass. Darker color represents larger error rate and is undesirable.}
    \label{thrustMappingRegion}
\end{figure}
\begin{figure*}
    \centering
    \includegraphics[width=\textwidth]{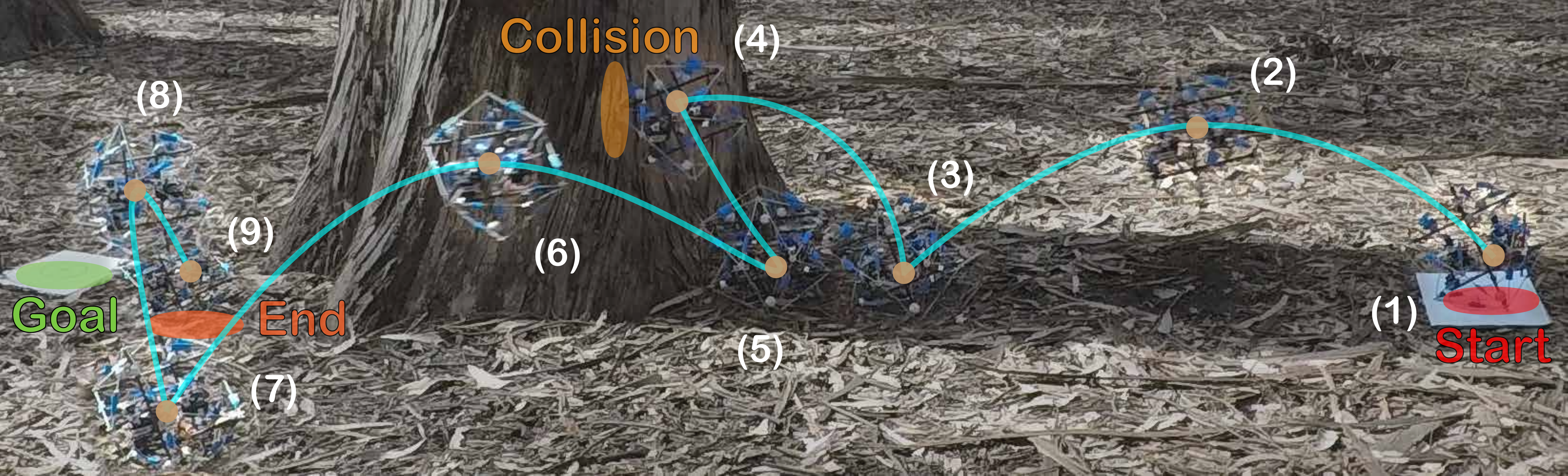}
    \caption{Composite image of the tensegrity aerial vehicle autonomously operating in a previously unknown forest environment. The cyan curve marks the movement of the vehicle. The vehicle is ordered to move from the start point on the right side of the figure to the goal point on the left. A tree obstacle exists between the two points. The vehicle successfully survives a collision with the tree and arrives at an endpoint close to the goal. The distance between the goal point and the end point is 0.25m. The background is desaturated to highlight the vehicle movement.}
    \label{ExploreForest}
\end{figure*}
We implemented and tested our re-orientation strategy in Section \ref{sec:modellingAndControl}. When generating the re-orientation paths, we assumed friction coefficient $\mu$ = 0.2, accounting for friction between the vehicle and slippery surfaces like wooden floors. The vehicle's ESCs were configured for bi-directional motor operation, thus providing increased torque generation authority during re-orientation. Fig. \ref{feasibleConnection}a illustrates feasible rotations between adjacent faces, while Fig. \ref{feasibleConnection}b presents the generated re-orientation paths. Notice that the vehicle can re-orient from any start face to the desired goal faces.

We also investigated the advantage of relaxing the constraint of sum of thrusts being zero in \cite{zha2020collision} via computing the additional payload that can be added to the center of mass before the re-orientation paths fail. We solved the optimization problems maximizing the vehicle mass under constraints \eqref{force_eq_edge} to \eqref{thrustrange} for all neighboring rotations. When an additional 0-thrust-sum constraint is imposed, the re-orientation paths will fail with an additional 12g mass. However, without this constraint, the vehicle can re-orient from all faces with an additional mass up to 30g.

In addition, we analyzed the advantage of the optimization-based torque-thrust converter in Section \ref{sec:modellingAndControl}.E. When thrust saturation occurs, the generated torque will deviate from the command. The rate of this error, defined as the ratio between the error's norm and the command's norm, is used to gauge the effectiveness of thrust conversion. To demonstrate the advantage of our method, we use the rotation from face 3 to face 4 as an example, as it requires a torque that rolls, pitches, and yaws the vehicle simultaneously.
We compare the error rates of three methods, as shown in Fig. \ref{thrustMappingRegion}: 
1) Adding an additional 0-thrust-sum constraint to the under-determined linear system Eq. \eqref{ThrustTorque} to make it fully-determined, solving the combined linear system for desired thrusts, and saturating the thrusts based on the feasible range.
2) Computing desired thrusts by solving Eq. \eqref{ThrustTorque} with the pseudoinverse method, and then saturating the thrusts. 
3) Computing the thrust commands by directly solving the optimization problems in \eqref{thrustMappingFeasible} and \eqref{thrustMappingInfeasible}. The figure shows that the optimization method has the largest error-free region. Furthermore, in scenarios with thrust saturation, the optimization-based method shows a much smaller error rate. This suggests that the optimization-based torque-thrust converter improves the vehicle's ability to implement re-orientation rotations.

With the planned re-orientation paths and the optimization-based thrust converter, the experimental vehicle can reliably re-orient and take off. Videos demonstrating successful re-orientations, including a scenario overcoming an initial failure caused by an external disturbance, are in the attachment.

\subsection{Autonomous operation in forest environment}
In this subsection, we present an experiment demonstrating the experimental vehicle's ability to autonomously operate in a cluttered environment. The vehicle is directed towards a goal in a forest previously unknown to the vehicle, which contains tree obstacles and uneven terrain.
We employ the Extended Kalman Filter (EKF) from \cite{mueller2017covariance} to estimate the vehicle's state, including position, velocity, and attitude.
Given the absence of external aids like GPS or motion capture, we employ the estimation strategy from \cite{wu2020using}, which improves estimation accuracy by breaking a long flight into short hops and updating the EKF with pseudo zero-velocity measurements when vehicle sensors indicate a stationary status.
Upon detecting a collision (the norm of accelerometer readings exceeds a threshold), the vehicle seeks to stabilize itself and land softly. After landing, it re-orients and attempts to hop around the obstacle it just encountered.

The outdoor environment experiment reveals certain limitations of the vehicle. Navigation accuracy is restricted by the inertial sensors' accuracy and range. High-impact collisions can cause sensor saturation, introducing significant error into the state estimator. Also, the re-orientation controller's performance can be hindered by torque limitations, particularly when the vehicle lands on steep slopes or is trapped by large ground indentations. To mitigate these issues, we lower the hop speed to avoid high-velocity collisions and instruct the vehicle to attempt backward hops when trapped.

Fig. \ref{ExploreForest} presents a composite image from the test. The vehicle was tasked to travel 3m in a specified direction with an unforeseen tree obstacle en route. During its second hop, the vehicle collided with the tree, managed to survive the impact, logged the obstacle's position, executed a sideways hop to evade the obstacle, and proceeded towards the goal. A video of this experiment is available in the attachment.
\section{Conclusion}
\label{sec:conclusion}
In this paper, we introduced the tensegrity aerial vehicle, a collision-resilient flying robot design with an icosahedron tensegrity structure. We established an approach for predicting structural stresses during collisions via a dynamics simulation, which facilitated component selection during the design process. This approach contributed to the successful creation of an experimental vehicle with strong collision resilience, capable of surviving a 7m drop with a 11.7m/s landing speed. Additionally, we developed a re-orientation controller, enabling the vehicle to take off post-collision. This combination of collision resilience and post-collision flight resumption makes the tensegrity aerial vehicle ideally suited for field operations in cluttered environments with hard-to-detect obstacles.

% Additionally, to mitigate the limitations of inertial sensors, we implemented an inertial navigation strategy that substitutes continuous flight with short hops. This allowed the vehicle to perform short-range autonomous operations without external sensors. 
\section*{Acknowledgements}
The experimental testbed at the HiPeRLab is the result of contributions of many people, a full list of which can be found at \href{https://hiperlab.berkeley.edu/members/}{\tt hiperlab.berkeley.edu/members/}. This work has been partially supported by the USDA AI Institute for Next Generation Food Systems (AIFS), USDA award number 2020-67021-32855, the Defense Advanced Research Projects Agency (DARPA) Subterranean Challenge and the UC Berkeley Fire Research Group. The authors would like to thank Alice Agogino, Alan Zhang and Douglas Hutchings for providing their insights on tensegrity design; Joey Kroeger, Natalia Perez and Bryan Yang for their help in the development of the tensegrity aerial vehicle; and reviewers for their valuable feedback.
%% BIBLIOGRAPHY
{
\bibliographystyle{IEEEtran}
\bibliography{bib/bibliography}

\begin{thebibliography}{10}
\providecommand{\url}[1]{#1}
\csname url@samestyle\endcsname
\providecommand{\newblock}{\relax}
\providecommand{\bibinfo}[2]{#2}
\providecommand{\BIBentrySTDinterwordspacing}{\spaceskip=0pt\relax}
\providecommand{\BIBentryALTinterwordstretchfactor}{4}
\providecommand{\BIBentryALTinterwordspacing}{\spaceskip=\fontdimen2\font plus
\BIBentryALTinterwordstretchfactor\fontdimen3\font minus
  \fontdimen4\font\relax}
\providecommand{\BIBforeignlanguage}[2]{{%
\expandafter\ifx\csname l@#1\endcsname\relax
\typeout{** WARNING: IEEEtran.bst: No hyphenation pattern has been}%
\typeout{** loaded for the language `#1'. Using the pattern for}%
\typeout{** the default language instead.}%
\else
\language=\csname l@#1\endcsname
\fi
#2}}
\providecommand{\BIBdecl}{\relax}
\BIBdecl

\bibitem{yasin2020unmanned}
J.~N. Yasin, S.~A.~S. Mohamed, M.-H. Haghbayan, J.~Heikkonen, H.~Tenhunen, and
  J.~Plosila, ``Unmanned aerial vehicles ({U}{A}{V}s): Collision avoidance
  systems and approaches,'' \emph{IEEE Access}, vol.~8, pp. 105\,139--105\,155,
  2020.

\bibitem{salaan2019development}
C.~J. Salaan, K.~Tadakuma, Y.~Okada, Y.~Sakai, K.~Ohno, and S.~Tadokoro,
  ``Development and experimental validation of aerial vehicle with passive
  rotating shell on each rotor,'' \emph{IEEE Robotics and Automation Letters},
  vol.~4, no.~3, pp. 2568--2575, 2019.

\bibitem{briod2014collision}
A.~Briod, P.~Kornatowski, J.-C. Zufferey, and D.~Floreano, ``A
  collision-resilient flying robot,'' \emph{Journal of Field Robotics},
  vol.~31, no.~4, pp. 496--509, 2014.

\bibitem{elios3}
Flyability, ``Elios 3,'' \url{https://www.flyability.com/elios-3}.

\bibitem{jia2022quadrotor}
H.~Jia, S.~Bai, R.~Ding, J.~Shu, Y.~Deng, B.~L. Khoo, and P.~Chirarattananon,
  ``A quadrotor with a passively reconfigurable airframe for hybrid terrestrial
  locomotion,'' \emph{IEEE/ASME Transactions on Mechatronics}, vol.~27, no.~6,
  pp. 4741--4751, 2022.

\bibitem{sareh2018rotorigami}
P.~Sareh, P.~Chermprayong, M.~Emmanuelli, H.~Nadeem, and M.~Kovac,
  ``Rotorigami: A rotary origami protective system for robotic rotorcraft,''
  \emph{Science Robotics}, vol.~3, no.~22, p. eaah5228, 2018.

\bibitem{de2021resilient}
P.~De~Petris, H.~Nguyen, M.~Kulkarni, F.~Mascarich, and K.~Alexis, ``Resilient
  collision-tolerant navigation in confined environments,'' in \emph{2021 IEEE
  International Conference on Robotics and Automation (ICRA)}.\hskip 1em plus
  0.5em minus 0.4em\relax IEEE, 2021, pp. 2286--2292.

\bibitem{mintchev2018bioinspired}
S.~Mintchev, J.~Shintake, and D.~Floreano, ``Bioinspired dual-stiffness
  origami,'' \emph{Science Robotics}, vol.~3, no.~20, p. eaau0275, 2018.

\bibitem{jang2019design}
J.~Jang, K.~Cho, and G.-H. Yang, ``Design and experimental study of
  dragonfly-inspired flexible blade to improve safety of drones,'' \emph{IEEE
  Robotics and Automation Letters}, vol.~4, no.~4, pp. 4200--4207, 2019.

\bibitem{shu2019quadrotor}
J.~Shu and P.~Chirarattananon, ``A quadrotor with an origami-inspired
  protective mechanism,'' \emph{IEEE Robotics and Automation Letters}, vol.~4,
  no.~4, pp. 3820--3827, 2019.

\bibitem{liu2021toward}
Z.~Liu and K.~Karydis, ``Toward impact-resilient quadrotor design, collision
  characterization and recovery control to sustain flight after collisions,''
  in \emph{2021 IEEE International Conference on Robotics and Automation
  (ICRA)}.\hskip 1em plus 0.5em minus 0.4em\relax IEEE, 2021, pp. 183--189.

\bibitem{de2022being}
R.~de~Azambuja, H.~Fouad, Y.~Bouteiller, C.~Sol, and G.~Beltrame, ``When being
  soft makes you tough: A collision-resilient quadcopter inspired by
  arthropods' exoskeletons,'' in \emph{2022 International Conference on
  Robotics and Automation (ICRA)}.\hskip 1em plus 0.5em minus 0.4em\relax IEEE,
  2022, pp. 7854--7860.

\bibitem{9351657}
H.~Zhou, A.~R. Plummer, and D.~Cleaver, ``Distributed actuation and control of
  a tensegrity-based morphing wing,'' \emph{IEEE/ASME Transactions on
  Mechatronics}, vol.~27, no.~1, pp. 34--45, 2022.

\bibitem{rimoli2016impact}
J.~J. Rimoli, ``On the impact tolerance of tensegrity-based planetary
  landers,'' in \emph{57th AIAA/ASCE/AHS/ASC Structures, Structural Dynamics,
  and Materials Conference}, 2016, p. 1511.

\bibitem{zhang2021orientation}
A.~Zhang, D.~Hutchings, M.~Gupta, and A.~Agogino, ``Orientation control of
  self-righting tensegrity landers,'' in \emph{International Design Engineering
  Technical Conferences and Computers and Information in Engineering
  Conference}, vol. 85451.\hskip 1em plus 0.5em minus 0.4em\relax American
  Society of Mechanical Engineers, 2021, p. V08BT08A025.

\bibitem{garanger2020soft}
K.~Garanger, I.~del Valle, M.~Rath, M.~Krajewski, U.~Raheja, M.~Pavone, and
  J.~J. Rimoli, ``Soft tensegrity systems for planetary landing and
  exploration,'' in \emph{Earth and Space 2021}, 2020, pp. 841--854.

\bibitem{sunspiral2013tensegrity}
V.~SunSpiral, G.~Gorospe, J.~Bruce, A.~Iscen, G.~Korbel, S.~Milam, A.~Agogino,
  and D.~Atkinson, ``Tensegrity based probes for planetary exploration: Entry,
  descent and landing (edl) and surface mobility analysis,''
  \emph{International Journal of Planetary Probes}, vol.~7, p.~13, 2013.

\bibitem{kim2016hopping}
K.~Kim, L.-H. Chen, B.~Cera, M.~Daly, E.~Zhu, J.~Despois, A.~K. Agogino,
  V.~SunSpiral, and A.~M. Agogino, ``Hopping and rolling locomotion with
  spherical tensegrity robots,'' in \emph{2016 IEEE/RSJ International
  Conference on Intelligent Robots and Systems (IROS)}.\hskip 1em plus 0.5em
  minus 0.4em\relax IEEE, 2016, pp. 4369--4376.

\bibitem{8460667}
S.~Mintchev, D.~Zappetti, J.~Willemin, and D.~Floreano, ``A soft robot for
  random exploration of terrestrial environments,'' in \emph{2018 IEEE
  International Conference on Robotics and Automation (ICRA)}, 2018, pp.
  7492--7497.

\bibitem{zappetti2022dual}
D.~Zappetti, Y.~Sun, M.~Gevers, S.~Mintchev, and D.~Floreano, ``Dual stiffness
  tensegrity platform for resilient robotics,'' \emph{Advanced Intelligent
  Systems}, vol.~4, no.~7, p. 2200025, 2022.

\bibitem{tensDrone}
H.~Cotgrove, ``Tensegrity drones,''
  \url{https://www.haydencotgrove.com/tensegrity-drones}, (Accessed on
  01/02/2022).

\bibitem{zha2020collision}
J.~Zha, X.~Wu, J.~Kroeger, N.~Perez, and M.~W. Mueller, ``A collision-resilient
  aerial vehicle with icosahedron tensegrity structure,'' in \emph{2020
  IEEE/RSJ International Conference on Intelligent Robots and Systems
  (IROS)}.\hskip 1em plus 0.5em minus 0.4em\relax IEEE, 2020, pp. 1407--1412.

\bibitem{savin2022mixed}
S.~Savin, A.~Al~Badr, D.~Devitt, R.~Fedorenko, and A.~Klimchik,
  ``Mixed-integer-based path and morphing planning for a tensegrity drone,''
  \emph{Applied Sciences}, vol.~12, no.~11, p. 5588, 2022.

\bibitem{wu2020using}
X.~Wu and M.~W. Mueller, ``Using multiple short hops for multicopter navigation
  with only inertial sensors,'' in \emph{2020 IEEE International Conference on
  Robotics and Automation (ICRA)}.\hskip 1em plus 0.5em minus 0.4em\relax IEEE,
  2020, pp. 8559--8565.

\bibitem{jessen1967orthogonal}
B.~Jessen, ``Orthogonal icosahedra,'' \emph{Nordisk Matematisk Tidskrift}, pp.
  90--96, 1967.

\bibitem{pellegrino1986matrix}
S.~Pellegrino and C.~R. Calladine, ``Matrix analysis of statically and
  kinematically indeterminate frameworks,'' \emph{International Journal of
  Solids and Structures}, vol.~22, no.~4, pp. 409--428, 1986.

\bibitem{beer2012mechanics}
F.~P. Beer, E.~R. Johnston, J.~T. DeWolf, and D.~F. Mazurek, \emph{Mechanics of
  Materials, Chapter 4-Pure Bending}, 6th~ed.\hskip 1em plus 0.5em minus
  0.4em\relax McGraw-Hill, 2012.

\bibitem{goyal2018dynamics}
R.~Goyal and R.~E. Skelton, ``Dynamics of class 1 tensegrity systems including
  cable mass,'' in \emph{Earth and Space 2018: Engineering for Extreme
  Environments}.\hskip 1em plus 0.5em minus 0.4em\relax ASCE, 2018, pp.
  868--876.

\bibitem{fenwick2000stiffness}
R.~Fenwick and D.~Bull, ``What is the stiffness of reinforced concrete walls,''
  \emph{SESOC journal}, vol.~13, no.~2, pp. 23--32, 2000.

\bibitem{SciPy2023}
{SciPy Developers}, ``scipy.integrate.radau,''
  \url{https://docs.scipy.org/doc/scipy/reference/generated/scipy.integrate.Radau.html}.

\bibitem{shuster1993survey}
M.~D. Shuster \emph{et~al.}, ``A survey of attitude representations,''
  \emph{Navigation}, vol.~8, no.~9, pp. 439--517, 1993.

\bibitem{mattingley2012cvxgen}
J.~Mattingley and S.~Boyd, ``Cvxgen: A code generator for embedded convex
  optimization,'' \emph{Optimization and Engineering}, vol.~13, no.~1, pp.
  1--27, 2012.

\bibitem{betafpv}
``{Pavo25 Frame Kit},'' \url{betafpv.com/products/pavo25-frame-kit}.

\bibitem{mueller2017covariance}
M.~W. Mueller, M.~Hehn, and R.~D’Andrea, ``Covariance correction step for
  kalman filtering with an attitude,'' \emph{Journal of Guidance, Control, and
  Dynamics}, vol.~40, no.~9, pp. 2301--2306, 2017.

\end{thebibliography}
}
\vspace{-40pt}
\begin{IEEEbiography}
[{\includegraphics[width=1in,height=1.25in,clip,keepaspectratio]{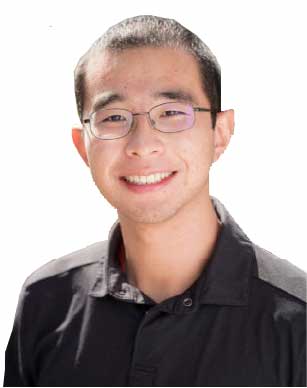}}]{Jiaming Zha}
received his bachelor of science degree from Rice University, USA in 2018 and master of science degree from University of California, Berkeley, USA in 2020.
He is currently a Ph.D. candidate at the High Performance Robotics Lab at UC Berkeley.
His research interests are novel mechanical design, control and path planning for aerial vehicles.
\end{IEEEbiography}
\vspace{-40pt}
\begin{IEEEbiography}
    [{\includegraphics[width=1in,height=1.25in,clip,keepaspectratio]{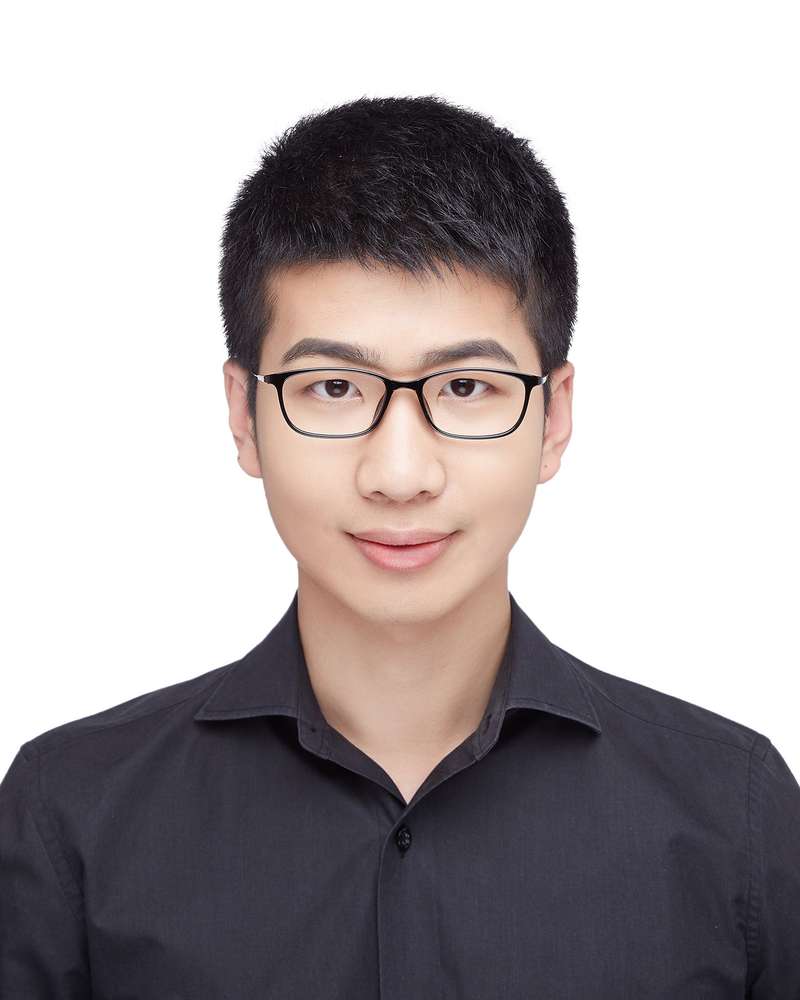}}]{Xiangyu Wu}
received his bachelor of science degree from Beijing Institute of Technology, China in 2017. He received his master of science degree in 2019 and his Ph.D. degress in 2022, both from University of California, Berkeley, USA. His research interests are the state estimation and path planning of multicopters.
\end{IEEEbiography}
\vspace{-40pt}
\begin{IEEEbiography}
    [{\includegraphics[width=1in,height=1.25in,clip,keepaspectratio]{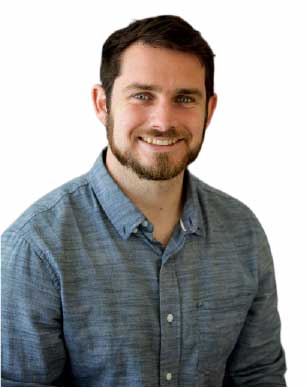}}]{Ryan Dimick}
received his bachelor of science degree from University of California, Davis, USA in 2017 and master of science degree from University of California, Berkeley, USA in 2021.
He is currently a robotics software engineer at Dusty Robotics.
\end{IEEEbiography}
\vspace{-40pt}
\begin{IEEEbiography}
    [{\includegraphics[width=1in,height=1.25in,clip,keepaspectratio]{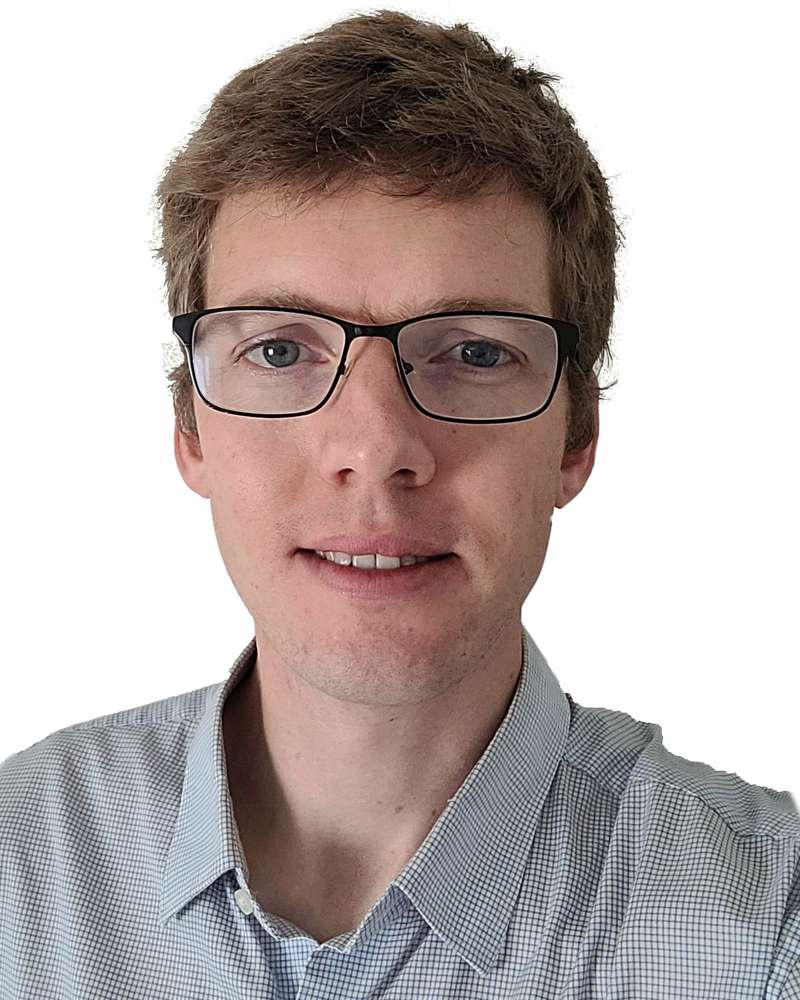}}]{Mark W. Mueller}
is an assistant professor of Mechanical Engineering at the University of California, Berkeley, and runs the High Performance Robotics Laboratory (HiPeRLab). He received a Dr.Sc. and M.Sc. from the ETH Zurich in 2015 and 2011, respectively, and a BSc from the University of Pretoria in 2008. His research interests include aerial robotics, their design and control, and especially the interactions between physical design and algorithms.
\end{IEEEbiography}
\end{document}